\def\year{2021}\relax
\documentclass[letterpaper]{article} % DO NOT CHANGE THIS
\usepackage{aaai21}  % DO NOT CHANGE THIS
\usepackage{times}  % DO NOT CHANGE THIS
\usepackage{helvet} % DO NOT CHANGE THIS
\usepackage{courier}  % DO NOT CHANGE THIS
\usepackage[hyphens]{url}  % DO NOT CHANGE THIS
\usepackage{graphicx} % DO NOT CHANGE THIS
\usepackage{natbib}  % DO NOT CHANGE THIS AND DO NOT ADD ANY OPTIONS TO IT
\usepackage{caption} % DO NOT CHANGE THIS AND DO NOT ADD ANY OPTIONS TO IT
\usepackage{epsfig}
\usepackage{color}
\usepackage{balance}  % for \balance command ON LAST PAGE  (only there!)
\usepackage{algorithmic}
\usepackage{algorithm}
\usepackage{multirow}
\usepackage{graphicx}
\usepackage{amsmath}
\usepackage{amssymb}
\usepackage{amsfonts}
\usepackage{xspace}
\usepackage{url}
\usepackage{bm}

\usepackage{color}
\usepackage[utf8]{inputenc}

\usepackage{amssymb}
\usepackage{amsbsy}
\usepackage{stmaryrd}
\usepackage{amsmath}
\usepackage{booktabs}
\usepackage{subcaption}
\newcommand{\B}{\boldsymbol}
\usepackage{bm}
\usepackage[switch]{lineno} 

\usepackage{color}

\DeclareMathOperator*{\argmin}{argmin}
\DeclareMathOperator*{\sbt}{s.t.}

\DeclareMathOperator*{\LR}{LR}
\DeclareMathOperator*{\ISE}{ISE}
\DeclareMathOperator*{\BIC}{BIC}

\newtheorem{prop}{Proposition}
\newtheorem{theorem}{Theorem}
\newtheorem{defn}{Definition}

\newtheorem{cor}{Corollary}

\newenvironment{myarray}[2][1]
{\def\arraystretch{#1}\array{#2}}
{\endarray}

\usepackage{tikz}
\usetikzlibrary{arrows,automata}

\makeatletter
    \let\@internalcite\cite
    \def\cite{\def\citename##1{##1}\@internalcite}
    \def\shortcite{\def\citename##1{}\@internalcite}
    \def\@biblabel#1{\def\citename##1{##1}[#1]\hfill}
\makeatother

\title{Sample-Efficient L0-L2 Constrained Structure Learning of Sparse Ising Models}
\author {
    Antoine Dedieu, Miguel Lázaro-Gredilla, Dileep George \\
}

\affiliations{
    Vicarious AI \\
    \{antoine, miguel, dileep\}@vicarious.com
}

\begin{document}
%\linenumbers 
\maketitle
\begin{abstract}
We consider the problem of learning the underlying graph of a sparse Ising model with $p$ nodes from $n$ i.i.d. samples. The most recent and best performing approaches combine an empirical loss (the logistic regression loss or the interaction screening loss) with a regularizer (an L1 penalty or an L1 constraint). This results in a convex problem that can be solved separately for each node of the graph. 
In this work, we leverage the cardinality constraint L0 norm, which is known to properly induce sparsity, and further combine it with an L2 norm to better model the non-zero coefficients. 
We show that our proposed estimators achieve an improved sample complexity, both (a) theoretically, by reaching new state-of-the-art upper bounds for recovery guarantees, and (b) empirically, by showing sharper phase transitions between poor and full recovery for graph topologies studied in the literature, when compared to their L1-based state-of-the-art methods.
\end{abstract}

% INTRODUCTION
\section{Introduction} \label{sec:introduction}

Ising models are extremely popular and useful tools \cite{onsager1944crystal,mccoy2012importance} with origins dating back to the $1920s$ \cite{ising1925beitrag} which have been widely studied in the fields of statistical physics \cite{thompson2015mathematical} and Bayesian modeling \cite{bishop2006pattern}.
Given an integer $p$ and a symmetric matrix $\bm{W}^* \in \mathbb{R}^{p \times p}$ with zero diagonal, a binary Ising graph (also known as fully visible Boltzman machine) with $p$ nodes and without external fields specifies the probability of a binary vector $\bm{z} \in \{ -1, 1\}^p$ as
\begin{equation}\label{ising}
p(\bm{z} | \bm{W}^*) = \frac{1}{Z (\bm{W}^*)} \exp\left( \frac{1}{2} \bm{z}^T \bm{W}^* \bm{z}  \right),
\end{equation}
where $Z(\bm{W}^*)$ is a normalization term called the partition function, which is defined as
\begin{equation}\label{partition}
\textstyle Z(\bm{W}^*) = \sum_{\bm{z} \in \{-1, 1\}^p} \exp\left( \frac{1}{2} \bm{z}^T \bm{W}^* \bm{z}  \right).
\end{equation}
We define the connectivity graph associated with $\bm{W}^*$ as the graph $G=(V,E)$ with vertices $V=\{1,\ldots,p\}$ and edges $E=\{(i,j): W^*_{ij} \ne 0 \}$. We additionally assume that each vertex of $G$ has a degree at most $k^*$ --- which is equivalent from saying that each row of $\B{W}^*$ has at most $k^*$ non-zero entries --- as it is traditionally the case in the literature \cite{bresler2015efficiently,vuffray2016interaction,lokhov2018optimal}. 

In this paper\footnote{Our code is available at \url{https://github.com/antoine-dedieu/structure_learning_sparse_ising_models}}, we consider the problem of learning the Ising model in Equation \eqref{ising} from $n$ samples independently drawn from the model. In particular, we are interested in analyzing the number of samples required for theoretical and empirical recovery of the connectivity graph. Maximum likelihood estimators are intractable for this task \cite{bishop2006pattern,welling2005learning} as the computational cost of estimating the partition function is exponential with the number of nodes in the graph. While early attempts to learn Ising models were based on mean-field approximation \cite{tanaka1998mean}, recent work has shown that the connectivity graph can be efficiently recovered by solving a convex problem for each node, without estimating the partition function. We review the two most recent and influential methods.

\medskip

\noindent
\textbf{Notation: }
For a vector $\bm{u}\in \mathbb{R}^p$ and an index $j\in \{1, \ldots,p\}$, we denote $\bm{u}_{-j}=(u_1, \ldots, u_{j-1}, u_{j+1}, u_p) \in \mathbb{R}^{p-1}$ the vector with all entries except the $j$th one.

\medskip

\noindent
\textbf{Method 1: Logistic regression estimators} For a sample $\bm{z}$ following Equation \eqref{ising} and a node $j \in \{1, \ldots, p\}$ of the Ising graph, the conditional probability of the $j$th observation $z_j$ given all the other observations $\bm{z}_{-j}$ can be expressed
\begin{equation}\label{gibbs}
p(z_j | \bm{z}_{-j}, \bm{W}^*) = \frac{1}{1 + \exp(-2 z_j (\bm{w}^*_{-j})^T \bm{z}_{-j}) },
\end{equation}
where we have noted $\bm{w}^*_{-j} = (\bm{e}_j^T \bm{W}^*)_{-j} \in \mathbb{R}^{p-1}$ the $j$th row of the connectivity matrix, without the (null) diagonal term. Consequently, given $n$ independent observations $\bm{z}^{(1)}, \ldots, \bm{z}^{(n)}$, the (normalized) pseudo-likelihood (PL) estimator \cite{besag1975statistical} of $\bm{W}^*$ is computed by solving the convex logistic regression (LR) problem for each node:
\begin{equation}\label{logreg}
\tilde{\bm{w}}_{-j} \in \argmin_{\bm{w} \in \mathbb{R}^{p-1}} \frac{1}{n} \sum_{i=1}^n \log(1 + \exp(-2 y_i \bm{x}_i^T \bm{w} )),
\end{equation}
where, for the $j$th node, we have we denoted $y_i=z^{(i)}_j$ and $\bm{x}_i = \B{z}^{(i)}_{-j}$ for the sake of simplicity. We can then derive an estimator $\tilde{\bm{W}}$ of $\bm{W}^*$ by solving Problem \eqref{logreg} for each node of the graph. The symmetrized estimator $\hat{\bm{W}}$ is then derived by sharing the parameters for both halves \cite{ravikumar2010high,vuffray2016interaction,lokhov2018optimal}, that is by defining $\hat{W}_{ij} = \frac{1}{2}(\tilde{W}_{ij} + \tilde{W}_{ji}), \forall i,j$. When the graph structure is known to be sparse, an influential work \cite{ravikumar2010high} proposes to add an L1 regularization to encourage sparsity in the coefficients. That is, the authors solve for each node the convex problem
\begin{equation}\label{l1-logreg}
\min_{\bm{w} \in \mathbb{R}^{p-1}} \frac{1}{n} \sum_{i=1}^n \log(1 + \exp(-2 y_i \bm{x}_i^T \bm{w} )) + \lambda \| \bm{w} \|_1,
\end{equation}
where the regularization parameter $\lambda\ge 0$ controls the degree of shrinkage on $\bm{w}$. A well-known limitation of their approach \cite{montanari2009graphical} is that their theoretical guarantees hold for a small class of models which satisfies restricted isometry properties (RIP) that are NP-hard to validate \cite{bandeira2013certifying}. To overcome this, a recent work \cite{wu2019sparse} considers the constrained version of Problem \eqref{l1-logreg} and estimates the sparse connectivity graph by solving:
\begin{equation}\label{l1-logreg-constrained}
\min_{\bm{w} \in \mathbb{R}^{p-1}} \frac{1}{n} \sum_{i=1}^n \log(1 + \exp(-2 y_i  \bm{x}_i^T \bm{w} )) ~~~ \sbt ~~~ \| \bm{w} \|_1 \le \lambda,
\end{equation}
where $\lambda$ is selected so that $\lambda \ge \max_{i=1, \ldots,n} \| \bm{w}^*_{-j} \|_1$. The authors additionally derive the best upper bound known on the number of samples required for theoretical graph recovery guarantees without any RIP (cf. Section \ref{sec:theory}).

\medskip

\noindent
\textbf{Method 2: Interaction screening estimator}
A recent work \cite{vuffray2016interaction} introduces as an alternative the L1-regularized interaction screening estimator (ISE), which is defined at node $j$ as a solution of the convex problem:
\begin{equation}\label{ise}
 \min \limits_{ \B{w} \in \mathbb{R}^{p-1} } \frac{1}{n} \sum_{i=1}^n \exp (- y_i \mathbf{x}_i^T  \B{w}) + \lambda \| \B{w} \|_1.
\end{equation}
The authors study the sample complexity of ISE for recovery guarantees (cf. Section \ref{sec:theory}) and additionally show that L1-regularized ISE outperforms L1-regularized LR through a rich variety of computational experiments (cf. Section. \ref{sec:results}).

\medskip

\noindent
\textbf{Estimating the graph structure via reestimation and hard-thresholding: }
Although the L1 regularization (and constraint) set some of the coefficients to $0$, they also retain a few entries with small absolute values. The following steps are used to improve the quality of the connectivity graph estimates returned by the LR and ISE procedures \cite{lokhov2018optimal,wu2019sparse}. 
\newline
\textbf{(1)} After selecting the edges connected to a node, the coefficients are reestimated by optimizing the unregularized (or unconstrained) objective only over these edges --- that is by solving Problems \eqref{l1-logreg} and \eqref{ise} with $\lambda=0$, and Problem \eqref{l1-logreg-constrained} with $\lambda=\infty$.
\newline
\textbf{(2)} After deriving the symmetrized graph estimate, the authors only keep the entries with absolute weights larger than $\eta / 2$, where $\eta$ is the minimum absolute edge weight of $\B{W}^*$:
\begin{equation}\label{def-min-edge}
\eta = \min \left\{ |W^*_{ij}|: ~ W^*_{ij} \ne 0 \right\}.
\end{equation}
Step (2) assumes the knowledge of $\eta$ in the optimization procedure. The authors do not derive an estimator of this minimum weight, nor do they propose to select it using a validation set. The explicit use of $\eta$ is then a shortcoming of all existing L1-based experiments. In contrast, we aim herein at deriving estimators of the connectivity graph which do not require the knowledge of $\eta$. 

\medskip

\noindent
\textbf{What this paper is about: }
Despite their good, well-studied, theoretical and empirical performance, both LR and ISE estimators rely on the use of a convex L1 regularization, and on the knowledge of $\eta$ (cf. Equation \eqref{def-min-edge}) to estimate sparse Ising graphs. In this work, we leverage insights from high-dimensional statistics \cite{lecture-notes,raskutti2011minimax} and propose two novel estimators that learn the graph structure of sparse Ising models by solving L0-L2 constrained versions of Problems \eqref{l1-logreg} and \eqref{ise}. The rest of this paper is organized as follows. In Section \ref{sec:estimator}, we define our estimators as solutions of non-convex L0-L2 constrained problems and propose a discrete first order algorithm with low computational cost and convergence guarantees to obtain high-quality solutions. In Section \ref{sec:theory}, we prove that our estimators achieve the best upper bounds known for theoretical recovery of the connectivity graph of sparse Ising models: our bounds improve over existing rates for L1-based procedures. Finally, Section \ref{sec:results} assesses the computational performance of our estimators through a variety of experiments, for the graph topologies studied in the literature.
%Section \ref{sec:theory} shows the appealing property of our estimator by deriving state-of-the-art bounds. Finally,  Section \ref{sec:results} assess its computational performance for Ising graph with $p\sim100$  nodes and $\sim 10^4$ coefficients.

\section{Cardinality Constrained Estimators}\label{sec:estimator}
We consider the Ising model defined in Equation \eqref{ising} where each node of the graph is of degree at most $k^*$. We aim at deriving sparse estimators of the connectivity matrix $\B{W}^*$ with better statistical performance than L1-based LR and ISE. Our first estimator extends the LR procedure, and estimates the $j$th row of the connectivity matrix $\B{w}^*_{-j}$ by solving the L0-L2 constrained logistic regression problem at node $j$:
\begin{equation}\label{l0-l2-logreg}
\begin{myarray}[1.3]{c c}
\min \limits_{\B{w}  \in \mathbb{R}^{p-1}} &   \frac{1}{n}  \sum \limits_{i=1}^n \log \left(1 + \exp (- 2y_i \mathbf{x}_i^T  \B{w}) \right) \\
\sbt &  \| \B{w}  \|_0 \le k, \ \| \B{w}  \|_2 \le \lambda / \sqrt{k},
\end{myarray}
\end{equation} 
where $y_i=z^{(i)}_j$ and $\bm{x}_i = \B{z}^{(i)}_{-j}$ as above.
Problem \eqref{l0-l2-logreg} minimizes the logistic loss with a constraint on the number of non-zeros of $\B{w}$ and a constraint on the L2 norm of $\B{w}$. Similarly, our second estimator extends the ISE procedure, and solves the L0-L2 constrained ISE problem, defined for the $j$th node as
\begin{equation}\label{l0-l2-ise}
\begin{myarray}{c c}
\min \limits_{\B{w}  \in \mathbb{R}^{p-1}} &   \frac{1}{n}  \sum \limits_{i=1}^n \exp (- y_i \mathbf{x}_i^T  \B{w}) \\
\sbt &  \| \B{w}  \|_0 \le k, \ \| \B{w}  \|_2 \le \lambda / \sqrt{k}. 
\end{myarray}
\end{equation} 
Problems \eqref{l1-logreg}, \eqref{l1-logreg-constrained} and \eqref{ise} consider one regularization parameter, $\lambda$, which simultaneously searches for a subset of features and shrinks the coefficients. In contrast, the L0 cardinality constraint in Problems  \eqref{l0-l2-logreg} and \eqref{l0-l2-ise} controls the model size while the L2 constraint controls the absolute values of the coefficients. The choice of the L2 constraint (over its L1 counterpart) is motivated by the theoretical guarantees of our estimators, which we present in Section \ref{sec:theory}.

\medskip

\noindent
\textbf{Solving L0-constrained problems: }
The use of the L0 constraint has been widely studied for the least-squares settings \cite{friedman2001elements}. Despite its excellent statistical properties \cite{bunea2007aggregation,raskutti2011minimax}, L0-constrained least-squares (aka \textit{best subsets}) has often been perceived as computationally infeasible \cite{natarajan1995sparse}. Consequently, L1 regularization (aka \textit{Lasso}) \cite{tibshirani1996regression} has often been proposed as a convex surrogate to promote sparsity, while achieving near-optimal statistical performance  \cite{bickel2009simultaneous}.  However, recent work has shown that mixed integer programming algorithms can solve L0-constrained regression and classification problems with $p \sim 10^6$ variables \cite{hazimeh2018fast,bertsimas2017sparse,dedieu2020learning} in times comparable to fast L1-based algorithms, while leading to large performance improvements. In addition, pairing L0 with an additional L1 or L2 penalty has been proved to drastically improve the performance of the standalone best subsets \cite{mazumder2017subset}. We leverage these results in designing our proposed estimators.

Despite a rich body of work for least-squares, the L0 penalty has never been used for recovering the graph structure of Ising models. We aim herein at bridging this gap by proposing the connectivity matrix estimates defined as solutions of Problems \eqref{l0-l2-logreg} and \eqref{l0-l2-ise}. In particular, we derive below a discrete first order algorithm that extends an existing framework and leverages the use of warm-starts to compute high-quality solutions for these non-convex problems.

\subsection{Discrete First Order Algorithm}\label{sec:dfo}
Inspired by proximal gradient methods for convex optimization \cite{nesterov2013gradient}, we propose a discrete first order (DFO) algorithm for obtaining good estimates of $\B{W}^*$ by solving the L0-L2 constrained Problems \eqref{l0-l2-logreg} and \eqref{l0-l2-ise}. Our procedure adapts an existing framework~\cite{bertsimas2016best}, has low computational complexity and uses warm-start to increase the quality of the solutions.

We assume that $f$ is a non-negative convex differentiable loss with $C$-Lipschitz continuous gradient, that is, it satisfies:
\begin{equation}\label{lip-grad-1}
\|\nabla f(\B{w}) - \nabla f(\B{v}) \|_{2} \leq C \| \B{w} - \B{v} \|_{2}~~~\forall~ \B{w}, \B{v} \in \mathbb{R}^{p-1}.
\end{equation}
When $f(\B{w}) = \frac{1}{n}  \sum_{i=1}^n \log \left(1 + \exp (- 2y_i \mathbf{x}_i^T  \B{w})\right)$ is the logistic loss, the Hessian of $f$ can be expressed as: 
$$\nabla^2 f(\B{w}) =\frac{1}{n} \sum_{i=1}^n  \frac{4\exp ( - 2 y_i \mathbf{x}_i^T  \B{w})}{ \left(1 + \exp ( - 2 y_i \mathbf{x}_i^T  \B{w} )\right)^2 } \mathbf{x}_i \mathbf{x}_i^T,$$
and we can use $C= n^{-1} \sigma_{\max}(\B{X}^T \B{X} )$, where we have noted $\B{X} \in \mathbb{R}^{n \times (p-1)}$ the matrix with $i$th row $\B{x}_i$, and where $\sigma_{\max}(.)$ is the maximum eigeinvalue of a matrix. Similarly, the Hessian of the interaction screening loss is:
$$\nabla^2 f(\B{w}) =\frac{1}{n} \sum_{i=1}^n  \exp ( - y_i\mathbf{x}_i^T  \B{w})\mathbf{x}_i \mathbf{x}_i^T,$$
for which we can use $C= n^{-1} e^{\lambda} \sigma_{\max}(\B{X}^T \B{X} )$. We describe a DFO method for the following problem:
\begin{equation}\label{gen-form-1}
\min f(\B{w}) ~~~ \sbt ~~~ \| \B{w} \|_{0} \leq k, ~~ \| \B{w} \|_{2} \le \theta,
\end{equation}
%\begin{align}\label{gen-form-1}
%\begin{split}
%\min \limits_{\B{w}  \in \mathbb{R}^{p-1}} & ~~~~~~~   f(\B{w}) \\
%& \sbt \| \B{w} \|_{0} \leq k, ~~ \| \B{w} \|_{2} \le \theta.
%\end{split}
%\end{align}
where $\theta \ge 0$. For $D\ge C$, we upper-bound the smooth $f$ around any $\B{v} \in \mathbb{R}^{p-1}$ with the quadratic form $Q_D(\B{v},.)$ defined $\forall \B{w} \in \mathbb{R}^{p-1}$ as
\begin{equation*}\label{convex-continuous-gradient}
f( \B{w} ) \le  Q_D(\B{v}, \B{w}) = f(\B{v}) + \nabla f(\B{v})^T(\B{w}-\B{v}) + \frac{D}{2} \| \B{w} - \B{v} \|_2^2.
\end{equation*}
Given a solution $\B{v}$, our method minimizes the upper-bound of $f$ around $\B{v}$. That is, it solves:
\begin{align} \label{key-proxy}
\begin{split}   
\hat{\B{w}} &\in \argmin_{ \B{w}: ~\|\B{w}\|_{0} \leq k, ~\|\B{w}\|_{2} \leq \theta }  Q_D(\B{v},\B{w}) 
\\ &\in \argmin_{ \B{w}: ~\|\B{w}\|_{0} \leq k, ~\|\B{w}\|_{2} \leq \theta } \left\| \B{w} - \left(\B{v} - \tfrac{1}{D} \nabla f(\B{v})  \right) \right\|_2^2.
\end{split} 
\end{align}
The above can be solved with the the proximal operator:
\begin{equation}\label{thresh-op-1}
\mathcal{S}(\bm{v} ; k; \theta ) := \argmin_{\B{w}: ~\|\B{w}\|_{0} \leq k, ~\|\B{w}\|_{2} \leq \theta}~~  \left\| \B{w} - \bm{v} \right\|_{2}^2,
\end{equation}
which can be computed with Proposition \ref{prop1}. The proof is presented in the Supplementary Materials. As the ordering of~$|v_{j}|$ may have ties, the solution of Problem \eqref{key-proxy} may be non-unique, and the solution of Problem \eqref{thresh-op-1} is set-based.

\begin{prop}\label{prop1}
Let $(1), \ldots, (p)$ be a permutation of the indices $1,\ldots, p$ such that the entries in $\bm{v}$ are sorted as:
$|v_{(1)}| \ge |v_{(2)}| \geq \ldots \ge |v_{(p)}|$. Then, $\hat{\B{w}} \in  \mathcal{S}(\bm{v} ; k; \theta )$ is given by:
\begin{equation*}
\hat{w}_{i} = \begin{cases} \min \left( 1, \frac{\theta}{\tau_{v}} \right) v_i & i \in \{ (1), (2), \ldots, (k) \} \\
0 & \text{otherwise},
\end{cases}
\end{equation*}
where $\tau_{v} =  \sqrt{\sum_{i=1}^{k} v_{(i)}^2}$ is the L2 norm of the~$k$  largest (absolute) entries of $\bm{v}$.
\end{prop}

\medskip

\noindent
\textbf{DFO algorithm: }
The DFO algorithm starts from an initialization $\B{w}^{(1)}$ and performs the following updates (for $t \geq 1$)
\begin{equation*}
\B{w}^{(t+1)}  \in  \mathcal{S} \left(  \B{w}^{(t)} - \tfrac{1}{D} \nabla f( \B{w}^{(t)}) ; k; \theta \right),
\end{equation*}
until some convergence criterion is met. In practice, we stop the algorithm when  $\| \B{w}^{(t+1)}  - \B{w}^{(t)} \|^2_{2} \leq \epsilon~$ for some small $\epsilon$ or when we have reached a maximum number $T_{\max}$ of iterations. Let $\tilde{\B{w}}$ denote the estimator returned and $\mathcal{I}(\tilde{\B{w}}) = \{i: ~ \tilde{w}_i \ne 0\}$ be the set of edges selected ---  which is of size at most $k$. The last step of the DFO algorithm reestimates the weights estimates of $\mathcal{I}(\tilde{\B{w}})$ by solving the convex problem
\begin{equation*}
\min f(\B{w}) ~~~ \sbt ~~~ w_i=0, ~ \forall i \notin \mathcal{I}(\tilde{\B{w}}).
\end{equation*}

\subsection{Convergence Properties}
We establish convergence properties of the sequence $\{\B{w}^{(t)}\}$ in terms of reaching a first order stationary point. Our work adapts the existing framework from~\citet{bertsimas2016best} to the constrained Problem \eqref{gen-form-1}. We first consider the following definition.

\begin{defn}\label{def-1} We say that $\B{w}$ is a
 first order stationary point of Problem~\eqref{gen-form-1} if
  $\B{w} \in \mathcal{S}( \B{w} - \frac{1}{L} \nabla f(\B{w}); k; \theta)$. We say that~$\B{w}$ is an $\epsilon$-accurate
 first order stationary point if  
 $\| \B{w} \|_{0} \leq k
 \text{ ; } \| \B{w} \|_{2} \leq \theta,$ and
 $\left\| \B{w} - \mathcal{S} \left( \B{w} - \frac{1}{D} \nabla f(\B{w}); k; \theta \right) \right\|^2_{2} \leq \epsilon.$
\end{defn}
We can now derive the following proposition. The proof is presented in the Supplementary Materials.
 \begin{prop}\label{prop-conv-1}
Let $\{\B{w}^{(t)}\}_{1\le t \le T}$  be a sequence generated by the DFO algorithm. Then, for $D > C$, the sequence $f(\B{w}^{(t)})$ is decreasing, and it converges to some~$f^* \ge 0$. In addition, the following convergence rate holds:
$$   \min_{1 \leq t \leq T - 1} \| \B{w}^{(t+1)}  - \B{w}^{(t)} \|_{2}^2 \leq \frac{2(f(\B{w}^{(1)} ) - f^*)}{T(D- C)}.$$
\end{prop}
%%A practical consequence of part (b) is that the support of the groups stabilizes after finitely many iterations. Once this happens,
%%Algorithm DFO ``works'' towards refining the values of~$\B\beta^{(m)}$, eventually leading to a stationary point of Problem~\eqref{grp-l0-1-const1}, which is given by
%%a solution to  $\min ~ g(\B\beta) ~\text{s.t.}~ \B\beta_{g} = 0, g \notin \texttt{S}({\B\beta}^{(m)})$. This ``polishing'' is precisely done by Step~2 of the DFO algorithm.
%Proposition~\ref{prop-conv-1} suggests that the DFO algorithm applied to Problem~\eqref{gen-form-1} leads to a decreasing sequence of bounded objective values, which then converges. 
Consequently, the algorithm reaches an $\epsilon$-accurate first order stationary point (cf. Definition~\ref{def-1}) in $O(\epsilon^{-1})$ iterations. In practice, the DFO algorithm converges much faster than the sublinear rate suggested by Proposition \ref{prop-conv-1}. This is the case when we leverage the use of warm-starts.

\subsection{Continuation Heuristic via Warm-starts}\label{sec:continuation}

Because Problem \eqref{gen-form-1} is non-convex, the DFO algorithm is sensitive to the initialization $\B{w}^{(1)}$. We design herein a continuation scheme which both (a) improves the quality of the solution and (b) decreases the computational cost when solving Problem \eqref{gen-form-1} at a particular value of $k$. We consider a decreasing list of parameters $\{k_1,\ldots, k_r\}$ with $k_1=p$ and $k_r=1$. For $k_1=p$, we solve Problem \eqref{gen-form-1} without the cardinality constraint\footnote{The value of $\theta$ can be tuned on an independent validation set.}. Our method iteratively refines the solution $\hat{\B{w}}(p)$ and returns a sequence of estimates $\{ \hat{\B{w}}(k_i) \}_{1\le i \le r}$ with decreasing support sizes. In particular, Problem \eqref{gen-form-1} at $k_r$ is solved by warm-starting with the solution obtained from $k_{r-1}$. We summarize the approach below.

\medskip

\noindent
\textbf{Continuation heuristic }
\newline
\textbf{(1)} Initialize $\hat{\B{w}}(k_1)$ by solving Problem \eqref{gen-form-1} without the cardinality constraint.
\newline
\textbf{(2)} For $i = 2, \ldots r$, set $\theta = 2 \| \hat{\B{w}}(k_{i-1}) \|_1$. 
\newline
Set $\hat{\B{w}}(k_i)$ as the output of the DFO algorithm initialized with $\hat{\B{w}}(k_{i-1})$, for $k_i$ and the above value of $\theta$.

\medskip

\noindent
\textbf{Specification to LR and ISE: }
The above heuristic can be applied to Problems \eqref{l0-l2-logreg} and \eqref{l0-l2-ise} to decrease the runtime of our DFO algorithm while returning higher-quality solutions. In practice we set $\hat{\B{w}}(p)$ as the respective solution of Problem \eqref{l1-logreg} for LR and Problem \eqref{ise} for ISE. In addition, the value of $k^*$ is unknown. As we return a sequence of estimators with decreasing support sizes, we propose to derive an estimator $\hat{k}$ of $k^*$ by minimizing the Bayesian information criterion \cite{schwarz1978estimating}. Our estimate $\hat{\B{w}}(\hat{k})$ will naturally be $\hat{k}$ sparse. Hence, contrary to all the L1-based approaches, we do not need the knowledge of $\eta$ (cf. Equation \eqref{def-min-edge}) to estimate the connectivity graph (cf. Section \ref{sec:results}).

\section{Statistical Properties}\label{sec:theory}

In this section, we study the statistical performance of the proposed L0-L2 constrained estimators for recovering the connectivity graph of the Ising model defined in Equation \eqref{ising}. In particular, we propose an upper bound on the number of samples required for high probability recovery. Our results do not require any external assumption and leverage the excellent statistical properties of the L0 constraint --- which are well-known for the sparse least-squares settings \cite{bunea2007aggregation,raskutti2011minimax,dedieu2020learning} --- to improve the best existing bounds, achieved by L1-based estimators. 

\medskip

\noindent
\textbf{Notations:}
We use the minimum edge weight $\eta$ defined in Equation \eqref{def-min-edge}. We additionally define the sparsity and width $(k^*, \lambda^*)$ of the connectivity matrix $\bm{W}^*$ as 
\vspace{-.3em}
$$\inf \left\{ (k, \lambda) ~ \bigg \rvert ~ \| \bm{w}^*_{-j} \|_0 \le k, ~ \sqrt{k} \| \bm{w}^*_{-j} \|_2 \le \lambda, ~ \forall j \right\}$$
where we have used the lexicographic ordering. In particular, it holds $\| \bm{w}^*_{-j} \|_1 \le \lambda^*, \forall j$.

\begin{table}[b!]
	\centering
	\begin{tabular}{p{0.15\textwidth}p{0.27\textwidth}}
	    \toprule
        Estimator + paper & Sample complexity \\
		\midrule
		Greedy method \cite{bresler2015efficiently} $\star$ & $O\left( \exp\left(\frac{\exp(O(k \lambda)) }{\eta^{O(1)}}\right) \log\left( \frac{p}{\delta} \right) \right)$ \\
		\midrule
		L1 ISE \cite{vuffray2016interaction} $\star$ & $O\left( \max \left( k, \frac{1}{\eta^2} \right) k^3 e^{6 \lambda} \log\left( \frac{p}{\delta} \right) \right)$ \\
		\midrule
		L1 LR \cite{lokhov2018optimal} $\star$ & $O\left( \max \left( k, \frac{1}{\eta^2} \right) k^3 e^{8 \lambda} \log\left( \frac{p}{\delta} \right) \right)$ \\
		\midrule
		L1-constrained LR \cite{lecture-notes} & $O\left(\frac{\lambda^2 \exp(8 \lambda)}{\eta^4} \log\left( \frac{p}{\delta} \right) \right)$ \\
		\midrule
		L1-constrained LR \cite{wu2019sparse} & $O\left(\frac{\lambda^2 \exp(12 \lambda)}{\eta^4} \log\left( \frac{p}{\delta} \right) \right)$ \\
		\midrule
		L0-L2 constrained LR (this paper) $\star$ &  $O\left(\frac{\lambda^2 \exp(8 \lambda)}{\eta^4} \log\left( \frac{p}{\sqrt{k}}\right) \log \left(\frac{2}{\delta} \right) \right)$ \\
		\midrule
		L0-L2 constrained ISE (this paper) $\star$ &  $O\left(\frac{\lambda^2 \vee \lambda^4 \exp(8 \lambda)}{\eta^4} \log\left( \frac{p}{\sqrt{k}}\right) \log \left(\frac{2}{\delta} \right) \right)$ \\
		\bottomrule
	\end{tabular}
	\caption{Comparison of the sample complexity required for graph recovery guarantees with probability at least $1 - \delta$, for an Ising model with $p$ nodes, width $\lambda^*$ and minimum absolute edge weight $\eta$. Part of the list is adapted from \citet{wu2019sparse}. Papers referred with a $\star$ additionally assume the degree of each node to be bounded by $k^*$. All approaches consider $\lambda \ge \lambda^*$ and $k\ge k^*$.}\label{table:bounds}
	\label{table:sota}
\end{table}

\subsection{Comparisons with Existing Work}
Table \ref{table:sota} compares the sample complexity of existing approaches for recovering sparse Ising graphs with high probability. Because $k^*$ and $\lambda^*$ are unknown, all  methods consider a couple $(k, \lambda)$ where $k\ge k^*$, and $\lambda \ge \lambda^*$. The table does not mention the L1-regularized LR estimator \cite{ravikumar2010high} defined as a solution of Problem \eqref{l1-logreg} as it relies on incoherence assumptions, which are NP-hard to validate \cite{bandeira2013certifying}. 
%In particular, the greedy estimator \cite{bresler2015efficiently} exhibits a doubly exponential dependency in the model width $\lambda$. This bound is improved by the L1-regularized ISE \cite{vuffray2016interaction} defined as a solution of Problem \eqref{ise}, which exhibits an exponential dependency in $\lambda$, and low-order polynomial dependency on $\lambda$ and $1 / \eta$. The authors later apply the same proof technique to the L1-regularized LR \cite{lokhov2018optimal}, reaching a similar bound. 
The L1-constrained LR \cite{lecture-notes,wu2019sparse} defined as a solution of Problem \eqref{l1-logreg-constrained} has been proved to reach the best upper bound known for sample complexity, and to be the most sample-efficient approach. Its bound scales exponentially with the width $\lambda$, logarithmically with the number of nodes $p$ and shows a polynomial dependency on $1 / \eta$. In this section, we prove that our proposed L0-L2 constrained estimators need fewer samples than all the existing approaches to recover the connectivity matrix with high probability. Our results are summarized in the last two rows of Table \ref{table:sota}.

\medskip

\noindent
\textbf{Connection with least-squares:}
The intuition behind the improved statistical performance of L0-L2 LR and L0-L2 ISE for recovering sparse Ising models can be derived from the least-squares problem. In this case, we consider the task of recovering a $k$ sparse vector in $\mathbb{R}^p$ from $n$ independent realizations of a Gaussian linear model. For mean squared error, best subsets achieves a rate of $O\left((k/n) \log(p/k) \right)$ which is known to be optimal \cite{ravikumar2010high,lecture-notes}. In contrast, Lasso achieves a rate of $O\left((k/n) \log(p) \right)$ under restricted eigenvalue assumptions \cite{bickel2009simultaneous}, which are NP-hard to satisfy.

\subsection{Upper Bound for L0-L2 Constrained LR}
The following theorem summarizes our main result for learning the connectivity matrix of an Ising model with the L0-L2 LR procedure. We refer to the Supplementary Materials for a detailed presentation and proof.
\begin{theorem} \label{upper-bound-logreg}
Let $\delta \in (0,1)$. The L0-L2 constrained logistic regression estimator of the connectivity matrix  $\hat{\B{W}}_{\LR}$, defined by solving Problem \eqref{l0-l2-logreg} for all nodes for the parameters $k \ge k^*, \lambda \ge \lambda^*$,  satisfies with probability at least $1 - \delta$:
\begin{equation*}
\| \hat{\B{W}}_{\LR} - \B{W}^* \|_{\infty}^2
\lesssim  \lambda e^{4 \lambda} \sqrt{\frac{\log(p / \sqrt{k}) }{n} \log(2 / \delta)}.
 \end{equation*}
\end{theorem}
\vspace{-.2em}
The constant in the bound is defined in the appendices.
We derive our upper bound on the number of samples required to recover with high probability the connectivity graph.
\begin{cor}
Let $\delta \in (0,1)$. Assume that the number of samples satisfies $n = O\left(\frac{\lambda^2 \exp(8 \lambda)}{\eta^4} \log\left( \frac{p}{\sqrt{k}}\right) \log\left( \frac{2}{\delta} \right) \right)$. The L0-L2 constrained logistic regression estimator derived by hard-thresholding $\hat{\B{W}}_{\LR}$ for the threshold $\eta /2$ recovers the exact connectivity graph with probability at least $1 -\delta$.
\end{cor}
When $k=k^*$ in Problem \eqref{l0-l2-logreg}, the L0-L2 LR estimator does not need the knowledge of $\eta$ to estimate the connectivity graph. In practice, we select $k$ to minimize some information criterion and do not use the thresholding step (cf. Section \ref{sec:results}).

\subsection{Upper Bound for L0-L2 Constrained ISE} 
Theorem \ref{upper-bound-ise} derives a similar upper bound than above when estimating the connectivity matrix of the Ising model in Equation \eqref{ising} with the L0-L2 ISE procedure.
\begin{theorem} \label{upper-bound-ise}
Let $\delta \in (0,1)$. The L0-L2 constrained interaction screening estimator of the connectivity matrix  $\hat{\B{W}}_{\ISE}$, defined by solving Problem \eqref{l0-l2-ise} for all nodes  for the parameters $k \ge k^*, \lambda \ge \lambda^*$  satisfies with probability at least $1 - \delta$:
\begin{equation*}
\| \hat{\B{W}}_{\ISE} - \B{W}^* \|_{\infty}^2
\lesssim (\lambda \vee \lambda^2) e^{4 \lambda} \sqrt{\frac{\log(p / \sqrt{k}) }{n} \log(2 / \delta)}.
 \end{equation*}
\end{theorem}
The sample complexity of the L0-L2 ISE follows.
\begin{cor}
Let $\delta \in (0,1)$. Assume that the number of samples satisfies $n = O\left(\frac{(\lambda^2 \vee \lambda^4) \exp(8 \lambda)}{\eta^4} \log\left( \frac{p}{\sqrt{k}}\right) \log\left( \frac{2}{\delta} \right) \right)$. The L0-L2 constrained interaction screening estimator derived by thresholding $\hat{\B{W}}_{\ISE}$ for the threshold $\eta /2$ recovers the exact connectivity graph with probability at least $1 -\delta$.
\end{cor}
As presented in Table \ref{table:sota}, both our proposed estimators achieve better upper bounds and are more sample-efficient than the existing approaches studied in the literature for learning Ising models. In particular, the dependency on the dimension size is lowered from a $\log(p)$ scaling to $\log(p/\sqrt{k})$. This is explained by the fact that the number of samples required for each node is lowered from a $\log(p)$ dependency to a $\log(p/k)$ one. The upper bounds then hold by applying a union bound on the $p$ nodes of the graph.

\medskip

\noindent
\textbf{Remark: } The decreasing dependency of our bounds with $k$ comes from the decreasing dependency of the L2 constraint with $k$ in Problems \eqref{l0-l2-logreg} and \eqref{l0-l2-ise}. This allows us to have a model width bounded by $\lambda$, as in the literature. We could alternatively use a fixed L2 constraint with value $\lambda$, and replace the model width bound by $\lambda \sqrt{k}$ for each method in Table \ref{table:sota}. Our bounds would now increase with $k$, while still improving over the existing ones for L1-based procedures.

\section{Computational Experiments}\label{sec:results} 
In this section, we assess the empirical performance of the proposed estimators for recovering the underlying graphs of various classes of Ising models. All the experiments were run on an Amazon Web Service c5.9 instance with 3.6GHz Xeon Platinum 8000 processor, 72GB of RAM.

\subsection{Data Generation}
We use the graph topologies that have been studied in the literature \cite{vuffray2016interaction,lokhov2018optimal} and consider the two following classes of connectivity graphs $G$:

\begin{figure}[htbp!]
    \centering
    \begin{tabular}{c c}
		\includegraphics[width=0.11\textwidth,height=0.07\textheight, clip = true ]{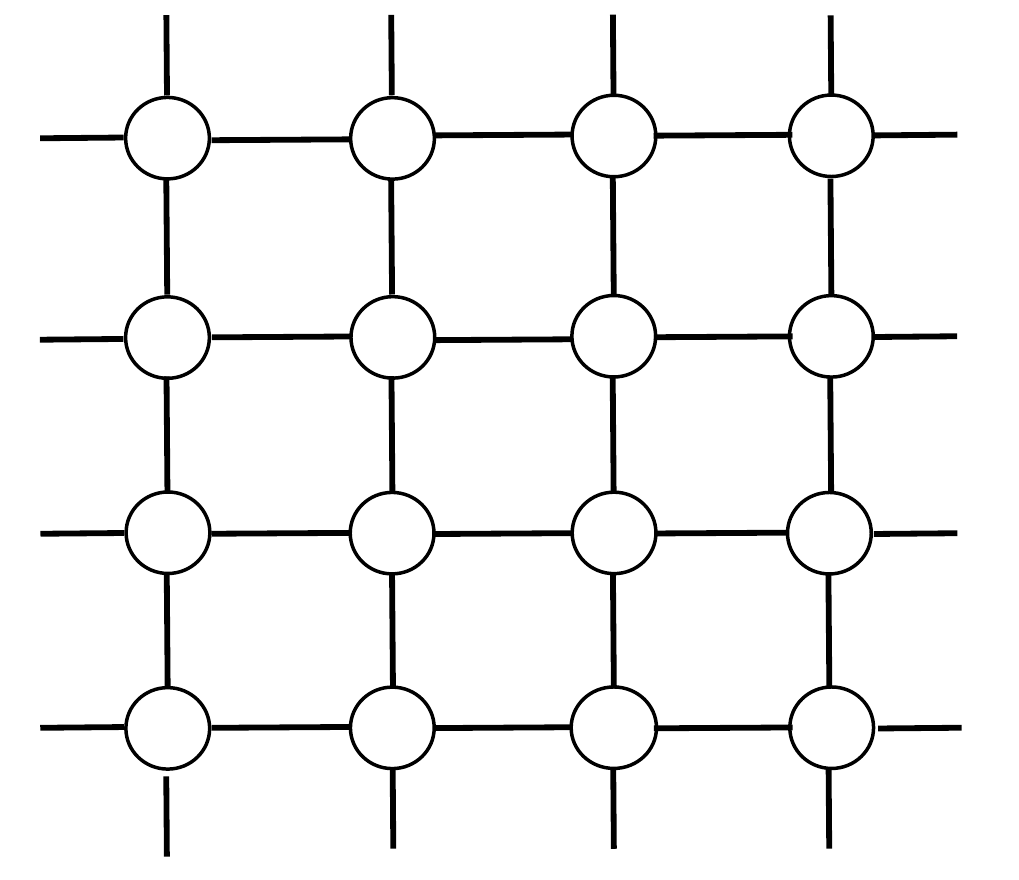}&
		\hspace{2em}
		\includegraphics[width=0.11\textwidth,height=0.07\textheight, clip = true ]{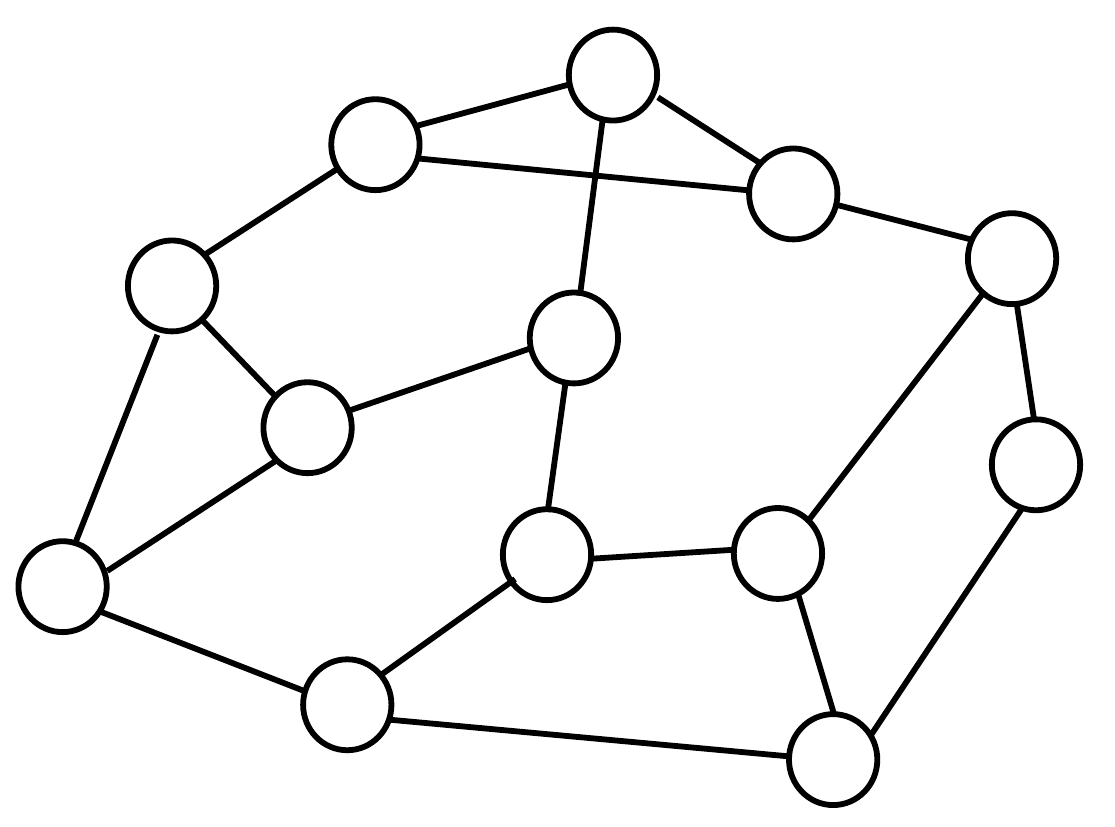}
	\end{tabular}
    \caption{Graph topologies for Examples 1 and 2}
    \label{fig:graphs}
\end{figure}
%\vspace{-1em}

\medskip

\noindent
\textbf{Example 1} We assume that $p$ is the square of an integer, and consider a four-connected two-dimensional lattice of size $\sqrt{p}$ with periodic boundary conditions. Each edge is of degree $k^*=4$. All the couplings take a similar value $\eta=0.5$\footnote{This setting has been proposed to compare LR with ISE \cite{lokhov2018optimal}. We decrease the value of $\eta$ from $0.7$ to $0.5$ to decrease the computational cost as (a) it does not affect the relative performance of the methods and (b) our experiments are more consuming than the authors as we tune the parameters.}.

\begin{figure*}[htbp!]
	\centering
		\begin{tabular}{c c}
		{\sf{Example 1, $p=16$}} &  {\sf{Example 1, $p=100$}} \\
		\includegraphics[width=0.48\textwidth,height=0.18\textheight, clip = true ]{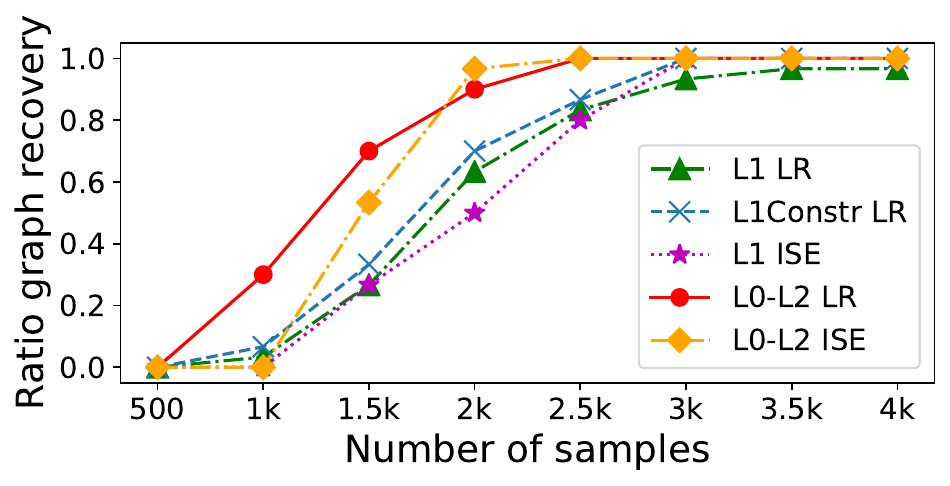}\label{fig:figa}&
		\includegraphics[width=0.48\textwidth,height=0.18\textheight, clip = true ]{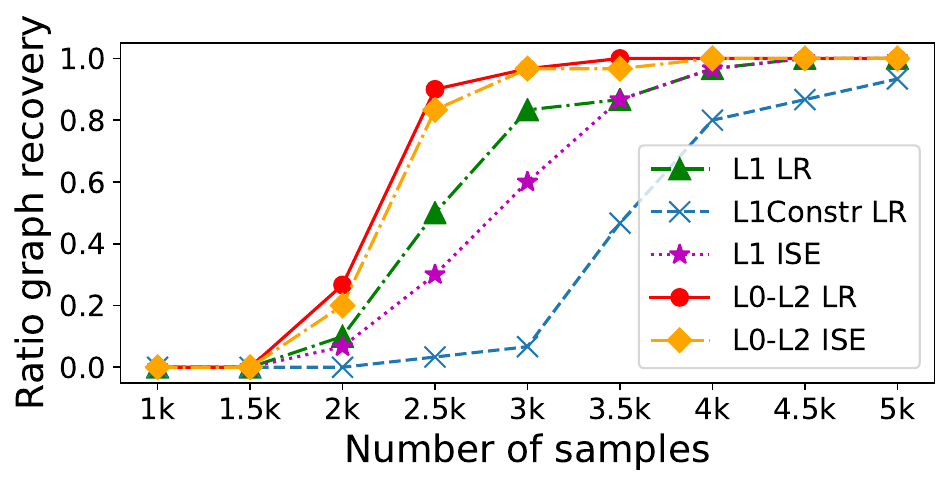}\label{fig:figb}
	\end{tabular}
	\smallskip
	\begin{tabular}{c c}
		{\sf{Example 2, $p=16$}} &  {\sf{Example 2, $p=100$}} \\
		\includegraphics[width=0.48\textwidth,height=0.18\textheight, clip = true ]{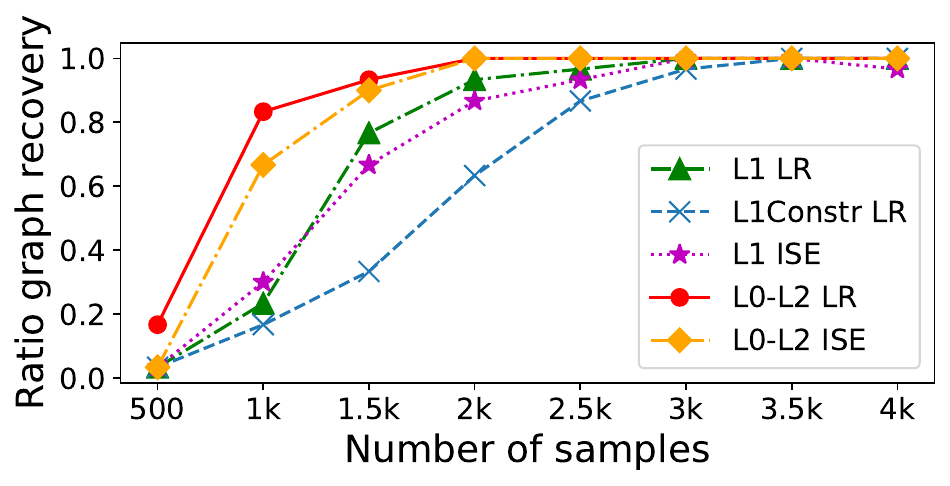}\label{fig:figc}&
		\includegraphics[width=0.48\textwidth,height=0.18\textheight, clip = true ]{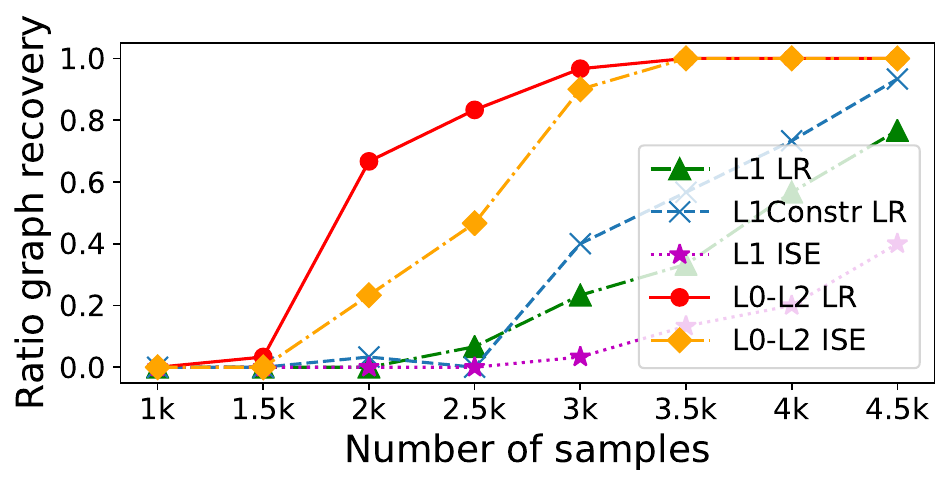}\label{fig:figd}
	\end{tabular}
	%{\sf Example1}
	\caption{
	Examples 1 and 2 with a small and a large number of nodes. The ratio of success is estimated over $30$ repetitions. Our proposed L0-L2 constrained estimators do not use the value of the minimum absolute edge weight $\eta$ to estimate the graph structure. Both outperform all their L1-based counterparts for both topologies while exhibiting sharper phase transitions.
	}
	\label{fig:first}
\end{figure*}

\medskip

\noindent
\textbf{Example 2:}  We consider a random regular graph with degree $k^*=3$. Couplings take random values that are uniformly generated in the range $[0.7, 0.9]$.

We simulate $n$ independent realizations from an Ising model with the above connectivity matrices. When $p\le16$, the partition function defined in Equation \eqref{partition} is computationally tractable and the samples are exactly drawn from the Ising model. When $p>16$, each observation is generated by running $1000$ iterations of a Gibbs sampler, exploiting the dynamics described in Equation \eqref{gibbs}.

\begin{figure*}[h!]
	\centering
    \includegraphics[width=0.6\textwidth,height=0.16\textheight, clip = true ]{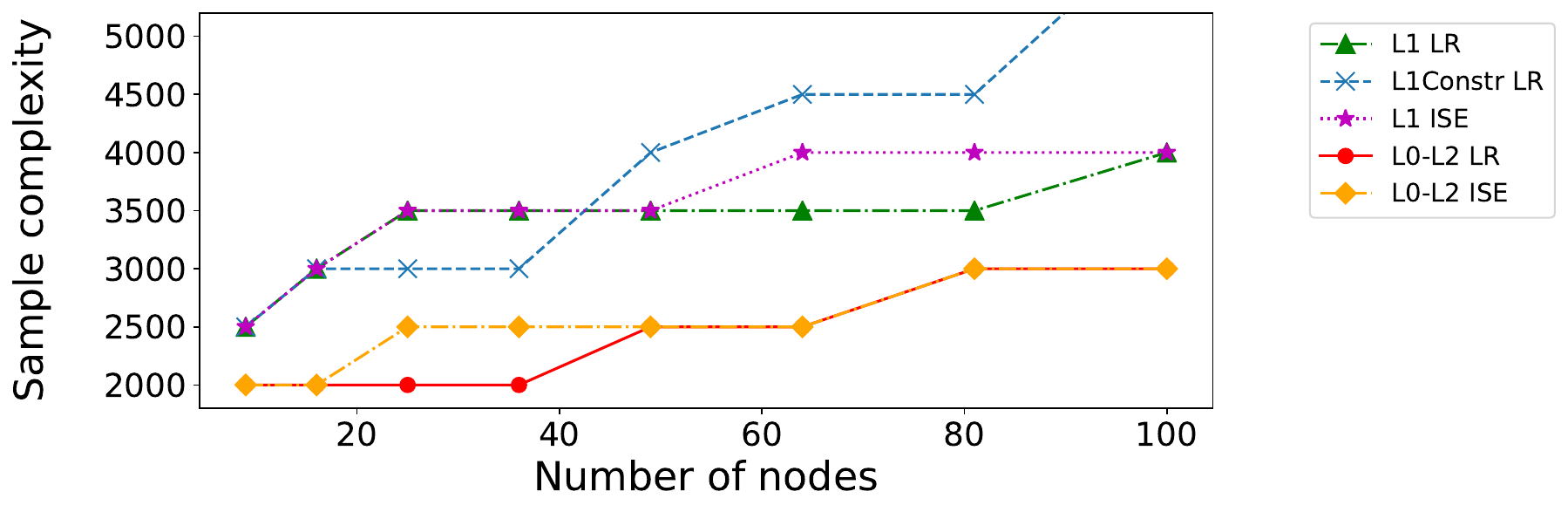}
	\caption{
	Example 1 with varying graph sizes. Both our L0-L2 constrained estimators constantly need fewer samples than their L1-based counterparts to fully recover the Ising graph structures. 
	}
	\label{fig:two}
\end{figure*}

\subsection{Methods Compared}
We compare the following methods in our experiments:

\medskip

$\bullet$ \textbf{L1 LR: } This is the L1-regularized logistic regression estimator \cite{ravikumar2010high} defined for each node as a solution of Problem \eqref{l1-logreg}, which we solve using Python's \texttt{scikit-learn} package \cite{scikit-learn}. For each node, we select the regularization parameter leading to the highest conditional likelihood on an independent validation set --- rather than setting it to the value suggested in \citet{lokhov2018optimal} --- as we observed that the former leads to better performance. More precisely, we compute a family of estimators for a decreasing geometric sequence of $20$ parameters $\lambda_1, \ldots, \lambda_{20}$ with common ratio $0.5$. We start from $\lambda_1 = 2 \| \B{X} \B{y} \|_{\infty}$ for which the solution of Problem \eqref{l1-logreg} is $\B{0}$.

\medskip

$\bullet$ \textbf{L1Constr LR: } This is the L1-constrained logistic regression estimator \cite{lecture-notes,wu2019sparse} defined for each node as a solution of Problem \eqref{l1-logreg-constrained}.
We implement a first order algorithm, using the FISTA acceleration procedure \cite{beck2009fast} for faster convergence. We use a stopping criterion $\epsilon =10^{-3}$ and a maximum number of $T_{\max}=300$ iterations.
Projection onto the L1 ball is achieved by the SPGL1 algorithm \cite{BergFriedlander:2008} using the software provided \cite{spgl1site}. 
We additionally tune the constraint parameter over the validation set.

\medskip

$\bullet$ \textbf{L1 ISE: } This is the L1-regularized interaction screening estimator \cite{vuffray2016interaction,lokhov2018optimal} defined for each node as a solution of Problem \eqref{ise}. Similarly to the above, we use an accelerated first-order algorithm and select, for each node, the regularization parameter leading to the highest conditional likelihood on the same validation set --- as we observed better empirical performance than using the value suggested in \citet{lokhov2018optimal}.

\medskip

$\bullet$ \textbf{L0-L2 LR: } This is the proposed L0-L2 constrained logistic regression estimator defined for each node as a solution of Problem \eqref{l0-l2-logreg}. We pair the DFO algorithm presented in Section \ref{sec:dfo} --- with a stopping criterion $\epsilon =10^{-3}$ and a maximum number of $T_{\max}=300$ iterations --- with the continuation heuristic described in Section \ref{sec:continuation}. For $k=p$, we initialize the heuristic with the L1 LR estimator. The value of $k$ is selected by minimizing the Bayesian information criterion (BIC) \cite{schwarz1978estimating} defined as $\BIC(k)= \log(n) S(k) - 2 \log(\mathcal{L})$ where $\mathcal{L}$ is the train pseudo-likelihood, $n$ the training set size and $S(k)$ the total number of edges of the estimated connectivity graph when the constraint is that each node is of degree at most $k$.

\medskip

$\bullet$ \textbf{L0-L2 ISE: } This is the proposed L0-L2 constrained interaction screening estimator, defined for each node as a solution of Problem \eqref{l0-l2-ise}. The computational and tuning procedures are similar to the ones for L0-L2 LR. We initialize the continuation heuristic for $k=p$ with L1 ISE.

\medskip

\noindent
\textbf{Estimating the graph structure: }
As presented in Section \ref{sec:introduction}, after solving each problem with the use of the corresponding regularization or constraint, we reestimate the objective without penalty only over the selected edges, and derive a symmetrized estimate by sharing the parameters for both halves. More importantly, because L0-L2 LR and L0-L2 ISE are naturally sparse, they do not need the knowledge of the minimum edge weight $\eta$ (cf. Equation \eqref{def-min-edge}). In contrast, for all the L1-based estimators, as described in their respective papers, we form the connectivity graph estimate by keeping the edges with absolute weights larger than $\eta / 2$.

\subsection{Phase Transition as $n$ Increases}\label{sec:first-exp}
Our first experiment compares the five different methods for Examples 1 and 2 and an increasing number of samples. More precisely, for a fixed value of $p$, the values of $n$ considered follow an increasing arithmetic sequence with common difference $500$. For each value of $n$, we simulate $30$ independent\footnote{Note that from the rule of three in statistics, the absence of failure of an event over $30$ repetitions guarantees a $95\%$ confidence interval where the probability of success is larger than $0.9$.} connectivity matrices $\B{W}^*$ and training sets of size $n$. We additionally generate a validation set of the same size than the training set, using the same $\B{W}^*$. For each run and each method, we select the best estimator as explained above and report the binary observation of whether the graph structure is fully recovered. We then compute the ratio of success for each method and each value of $n$ over the $30$ realizations.

We report our findings in Figure \ref{fig:first}. We consider Examples 1 and 2 for a small number of nodes $p=16$ %(for which the data exactly comes from the Ising model) 
and a large number of nodes $p=100$. %(for which we use a Gibbs-sampling procedure). 
All the methods in Figure \ref{fig:first} show a phase transition: their performance increases as the number of samples increases.
In addition, for both examples and both sizes, our two proposed L0-L2 constrained estimators show better empirical performance and sharper phase transitions than all the L1-based procedures. In particular, our methods both need fewer samples than their counterparts to exactly recover the Ising graph structures.

We complement our findings by reporting the L2 norm of the difference between $\B{W}^*$ and the connectivity matrix estimate of each method (before thresholding for L1-based ones) in the Supplementary Materials. While this metric --- referred to as L2 estimation --- does not need the knowledge of $\eta$ (for L1-based procedures), our estimators still show important gains in performance.
%\AD{We empirically observe that when we increase the correlation between neighboring variables (i.e. when we increase we absolute value of the entries of $W^*$), the number of samples required to recover the graph structure increases. However the relative performance of the methods is unchanged.}

\subsection{Performance Gain for All Graph Sizes}
The above suggests that, for $p\in \{16, 100\}$, our L0-L2 constrained estimators need fewer samples than L1-based methods to recover sparse Ising graphs. We explore herein whether this property holds for all graph sizes. For an estimator and a graph size $p$, the sample complexity $m^*(p)$ has been defined \cite{vuffray2016interaction,lokhov2018optimal} as the smallest number of samples the estimator requires to learn the connectivity graph of a model with $p$ nodes for all runs. We propose to use the more robust metric where we allow at most three failures (e.g. $90\%$ success) over the $30$ repetitions. We additionally define the sample complexity over all graph sizes smaller than $p$ as $n^*(p) = \max_{q \le p} m^*(q)$.

We consider Example 1 with the sequence of $p$ values $\{9, 16, 25, 36, 49, 64, 81, 100\}$ and run the same experiments as in Section \ref{sec:first-exp}. For each method and each value of $p$, we compute the corresponding value of $n^*(p)$ defined above. We report our findings in Figure \ref{fig:two}. For every graph size, our proposed L0-L2 constrained estimators both outperform all the L1-based procedures. That is, our approaches need fewer samples for recovering the Ising graphs studied in \citet{vuffray2016interaction,lokhov2018optimal} than the best known methods. In addition, as suggested by our theoretical results, each estimator needs more samples to recover larger graphs. We also note that (a) L1Constr LR performance decrease when $p$ increases and (b) the validation procedure does not give a winner between L1 LR and L1 ISE contrary to what has been reported with a fixed theoretically-driven parameter. Finally, we show in the Supplementary Materials that, for every value of $p$, our estimators achieve sharper phase transitions and important gains for L2 estimation.

\section{Further Work}
Although this paper focuses on learning binary Ising models, an interesting follow-up work would extend the proposed estimators for learning general discrete graphical models \cite{ravikumar2010high,wu2019sparse}.

%\newpage
%\section*{Ethics statement}
%The theoretical and empirical study of sparse statistical learning estimators in machine learning aims at (a) providing a rigorous understanding of the properties of statistical procedures underlying many modern applications of the field and (b) efficiently building interpretable models which deliver actionable insight for the decision-maker. As with any technology, negative consequences are possible but difficult to predict at this time. This is not a deployed system with immediate failure consequences or that can by itself leverage harmful biases, although these are possible when integrated into more complex systems.

\section*{Acknowledgements}
We thank Guangyao Zhou for his helpful comments on the manuscript.
\bibliography{aaai}

\end{document}

\maketitle
\section{Additional experiments}
We present herein some additional computational experiments which complement the findings reported in the main paper.

\subsection{L2 norm difference for Figure 2}
First, we introduce a new metric, which we refer below as \textit{L2 estimation}. It computes the L2 norm of the difference $\| \hat{\B{W}}^{(m)} - \B{W}^* \|_2$ between the estimator of the connectivity matrix returned by the  $m$th method and the ground truth $\B{W}^*$. Note that this method does not require the use of the minimum absolute edge weight $\eta$ of the connectivity graph.

\medskip
\noindent
The next Figure \ref{fig:first-appendix} complements Figure 2 in the main paper and shows the L2 estimation performance for the experiments run in Section 4.3. We average the metric over the $30$ runs and report the standard deviations. For Examples 1 and 2 and both graph sizes, both our L0-L2 constrained estimators achieve important gain for L2 estimation compared to all the L1-based procedures.

\begin{figure*}[htbp!]
	\centering
		\begin{tabular}{c c}
		{\sf{Example 1, $p=16$}} &  {\sf{Example 1, $p=100$}} \\
		\includegraphics[width=0.48\textwidth,height=0.18\textheight, clip = true ]{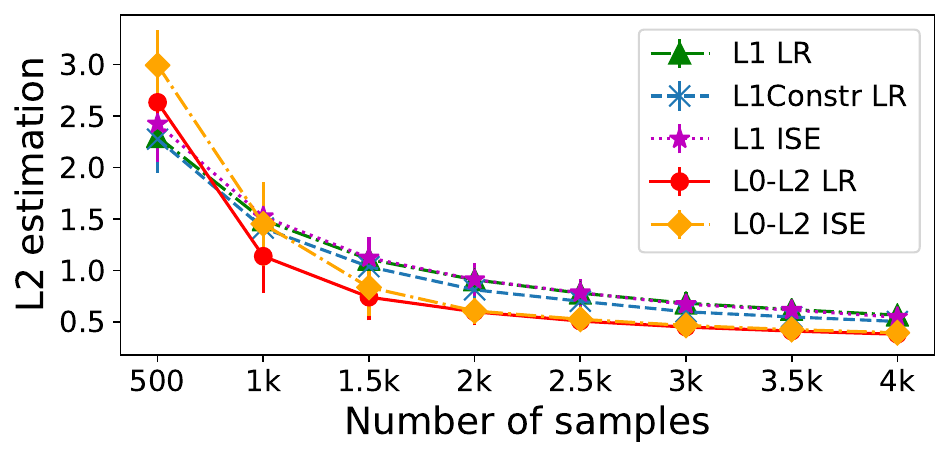}\label{fig:figa}&
		\includegraphics[width=0.48\textwidth,height=0.18\textheight, clip = true ]{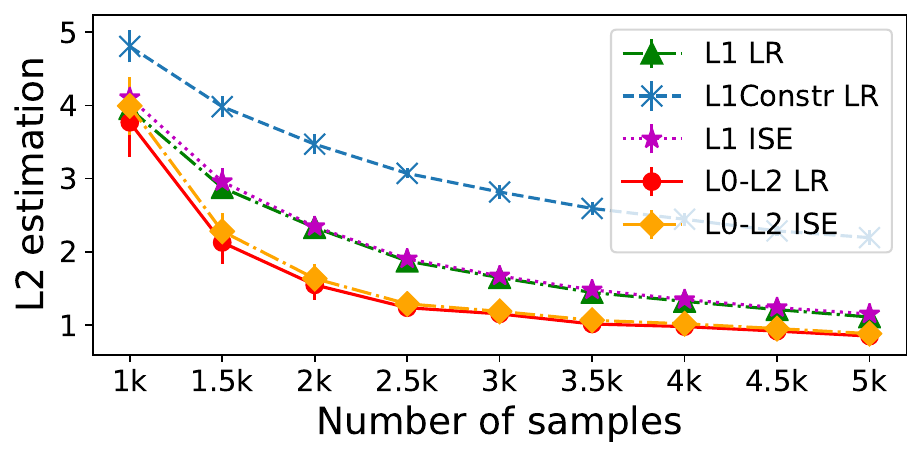}\label{fig:figb}
	\end{tabular}
	\smallskip
	\begin{tabular}{c c}
		{\sf{Example 2, $p=16$}} &  {\sf{Example 2, $p=100$}} \\
		\includegraphics[width=0.48\textwidth,height=0.18\textheight, clip = true ]{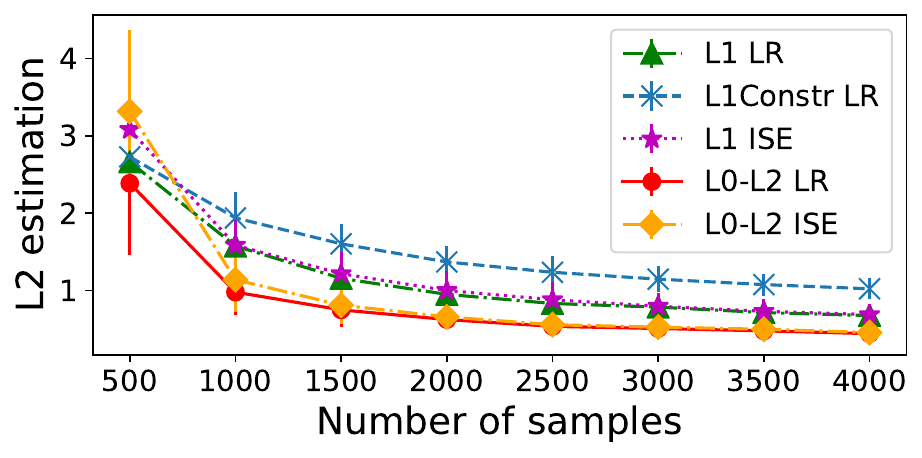}\label{fig:figc}&
		\includegraphics[width=0.48\textwidth,height=0.18\textheight, clip = true ]{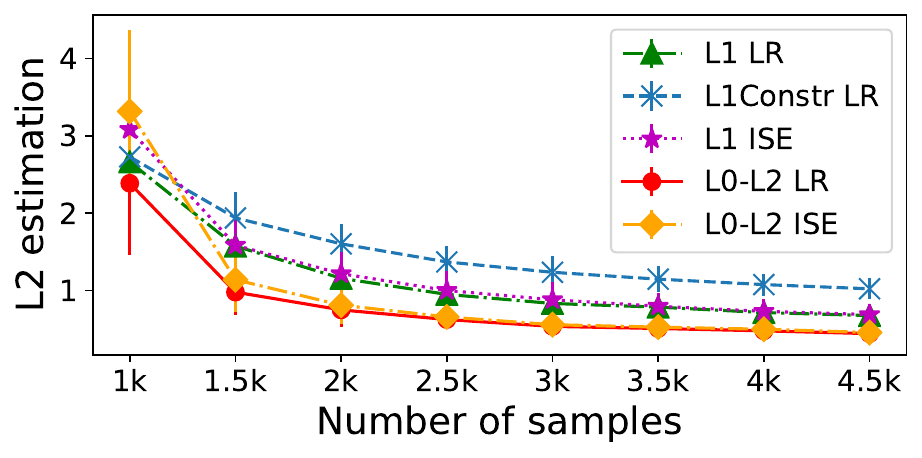}\label{fig:figd}
	\end{tabular}
	%{\sf Example1}
	\caption{\small{ 
	L0-L2 LR and ISE perform better for L2 estimation for Examples 1 and 2 and both graph sizes.
	}}
	\label{fig:first-appendix}
\end{figure*}

\subsection{Additional experiments for Figure 3}
Our next figure presents the phase transitions for all the values of $p$ used to derive Figure 3 in the main paper. We observe that for every graph size, our estimators show sharper phase transitions. 
\begin{figure*}[h!]
	\centering
	\begin{tabular}{c c}
		{\sf{Example 1, $p=9$}} &  {\sf{Example 1, $p=16$}} \\
		\includegraphics[width=0.48\textwidth,height=0.18\textheight, clip = true ]{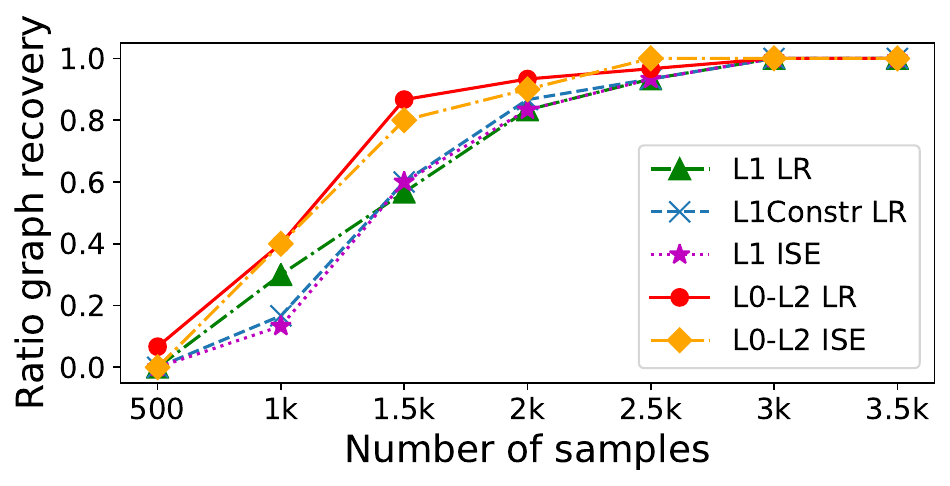}\label{fig:figa}&
		\includegraphics[width=0.48\textwidth,height=0.18\textheight, clip = true ]{periodic_graph_uniform_sign_P_16_rho_05.pdf}\label{fig:figb}
	\end{tabular}
	\smallskip
	\begin{tabular}{c c}
		{\sf{Example 1, $p=25$}} &  {\sf{Example 1, $p=36$}} \\
		\includegraphics[width=0.48\textwidth,height=0.18\textheight, clip = true ]{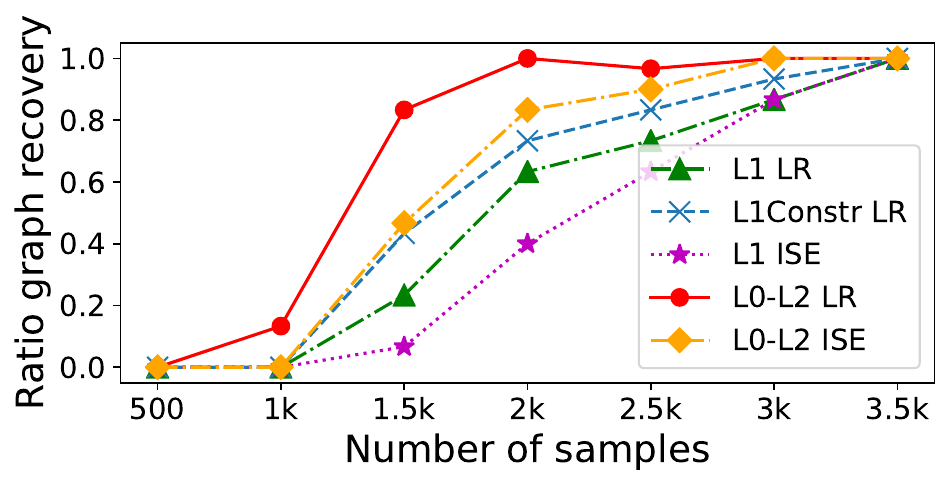}\label{fig:figa}&
		\includegraphics[width=0.48\textwidth,height=0.18\textheight, clip = true ]{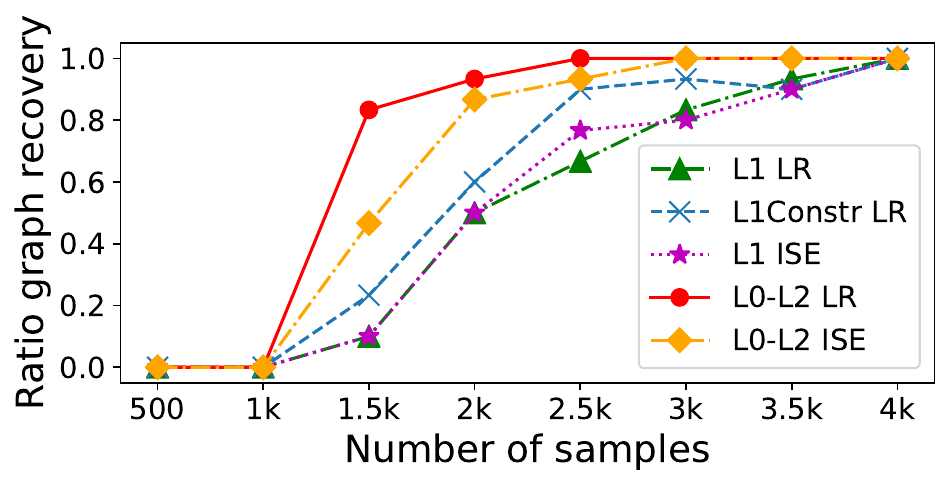}\label{fig:figb}
	\end{tabular}
	\smallskip
	\begin{tabular}{c c}
		{\sf{Example 1, $p=49$}} &  {\sf{Example 1, $p=64$}} \\
		\includegraphics[width=0.48\textwidth,height=0.18\textheight, clip = true ]{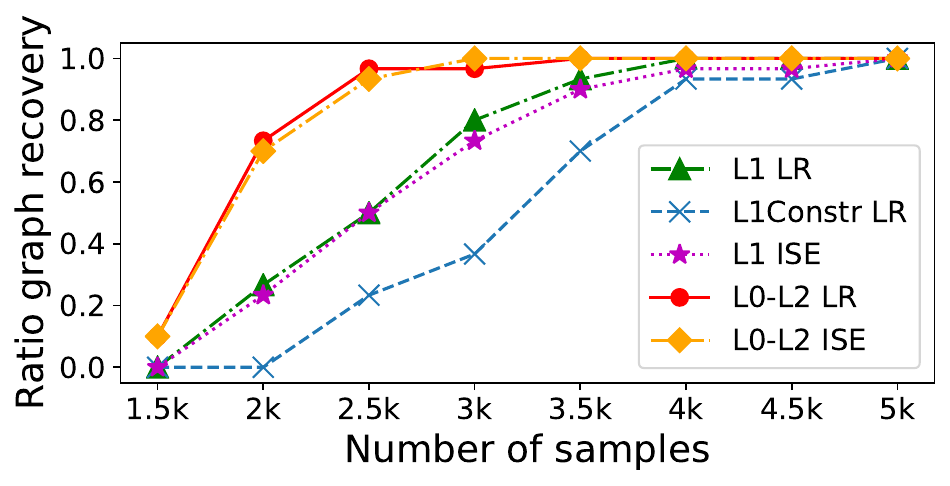}\label{fig:figa}&
		\includegraphics[width=0.48\textwidth,height=0.18\textheight, clip = true ]{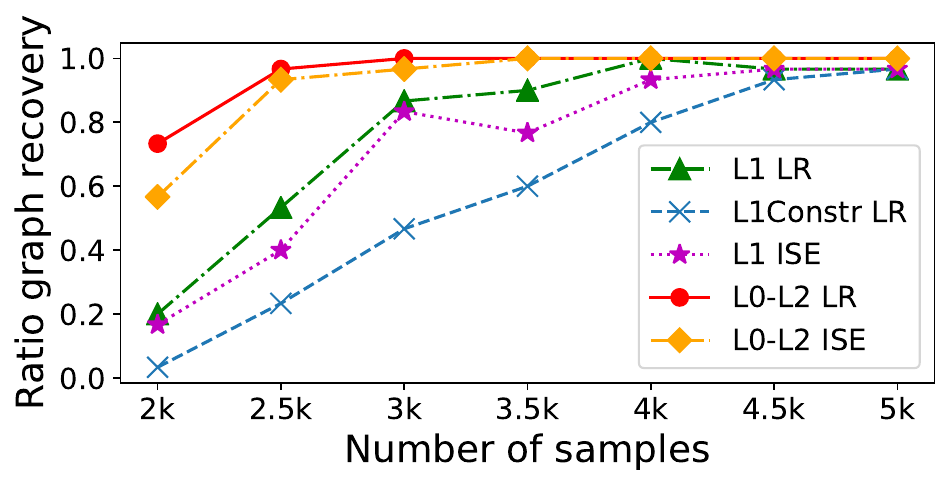}\label{fig:figb}
	\end{tabular}
	\smallskip
	\begin{tabular}{c c}
		{\sf{Example 1, $p=81$}} &  {\sf{Example 1, $p=100$}} \\
		\includegraphics[width=0.48\textwidth,height=0.18\textheight, clip = true ]{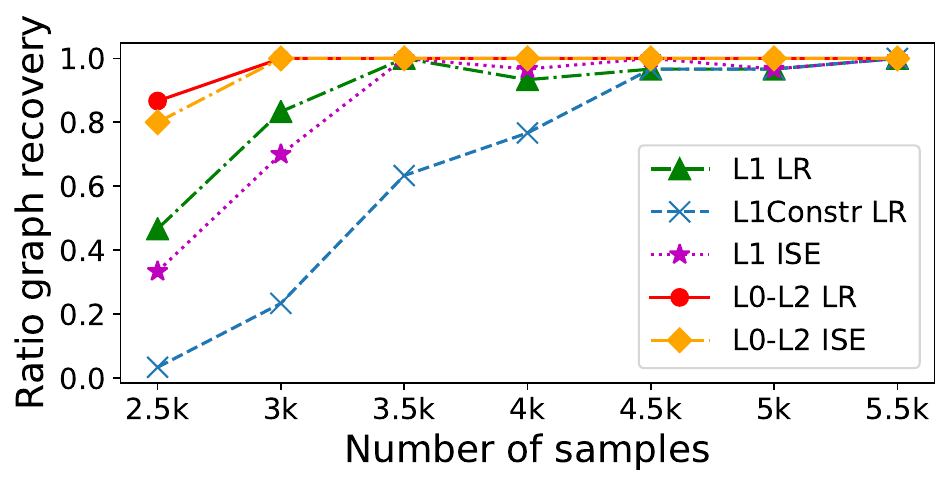}\label{fig:figa}&
		\includegraphics[width=0.48\textwidth,height=0.18\textheight, clip = true ]{periodic_graph_uniform_sign_P_100_rho_05.pdf}\label{fig:figb}
	\end{tabular}
	%{\sf Example1}
	\caption{\small{ 
	Our proposed methods constantly exhibit sharper phase transitions than the state-of-the-art L1-based approaches. Note that the plots for $p=16$ and $p=100$ appear in the main paper.
	}}
	\label{fig:first}
\end{figure*}

\newpage
We additionally report the corresponding performance for L2 estimation. 
\begin{figure*}[h!]
	\centering
	\begin{tabular}{c c}
		{\sf{Example 1, $p=9$}} &  {\sf{Example 1, $p=16$}} \\
		\includegraphics[width=0.48\textwidth,height=0.18\textheight, clip = true ]{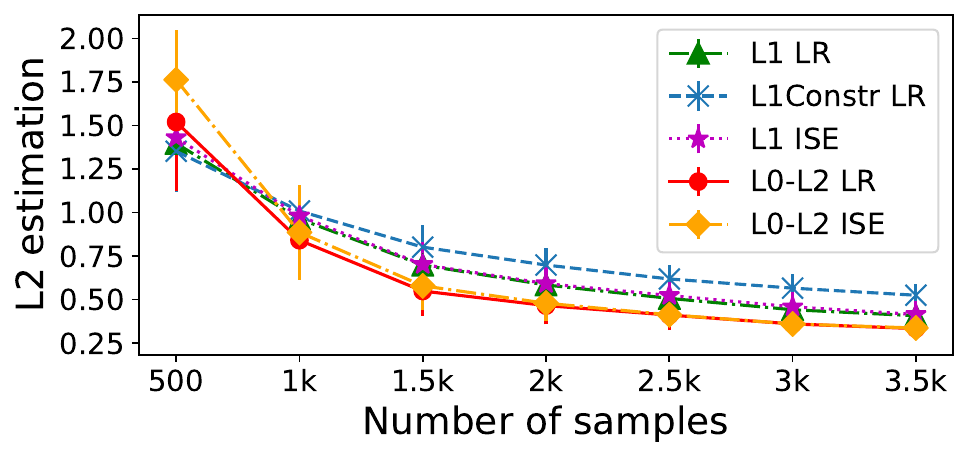}\label{fig:figa}&
		\includegraphics[width=0.48\textwidth,height=0.18\textheight, clip = true ]{periodic_graph_uniform_sign_P_16_rho_05_L2norm.pdf}\label{fig:figb}
	\end{tabular}
	\smallskip
	\begin{tabular}{c c}
		{\sf{Example 1, $p=25$}} &  {\sf{Example 1, $p=36$}} \\
		\includegraphics[width=0.48\textwidth,height=0.18\textheight, clip = true ]{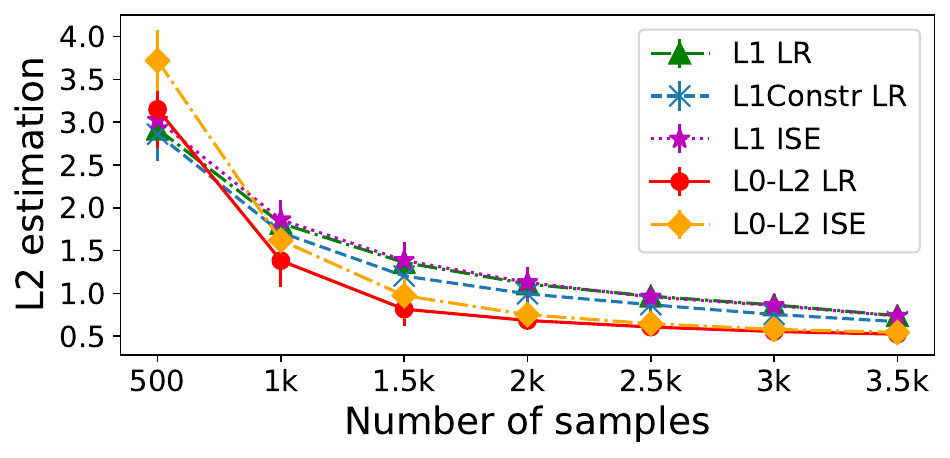}\label{fig:figa}&
		\includegraphics[width=0.48\textwidth,height=0.18\textheight, clip = true ]{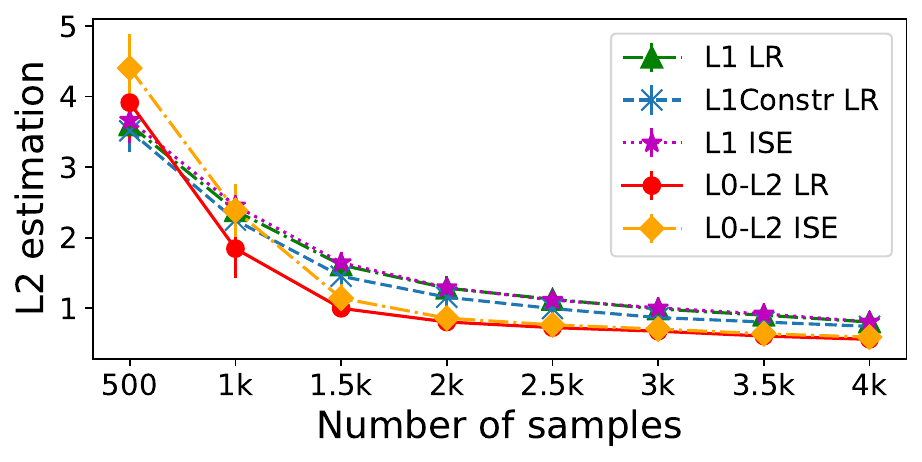}\label{fig:figb}
	\end{tabular}
	\smallskip
	\begin{tabular}{c c}
		{\sf{Example 1, $p=49$}} &  {\sf{Example 1, $p=64$}} \\
		\includegraphics[width=0.48\textwidth,height=0.18\textheight, clip = true ]{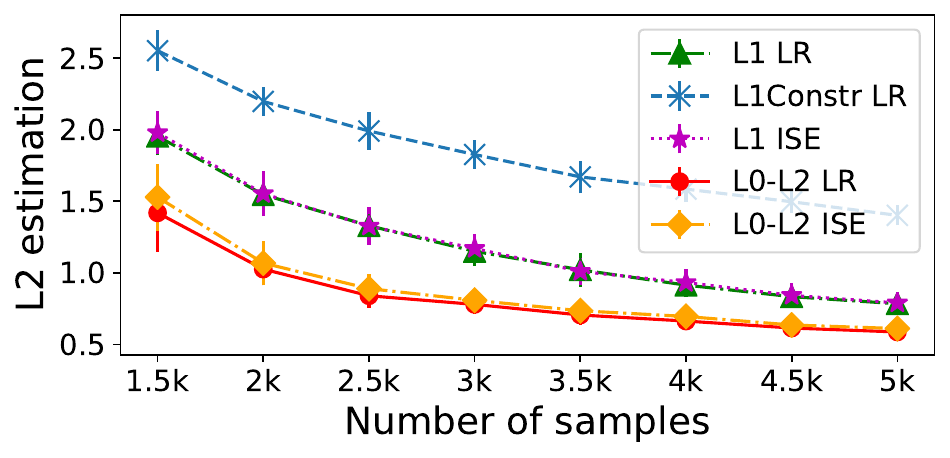}\label{fig:figa}&
		\includegraphics[width=0.48\textwidth,height=0.18\textheight, clip = true ]{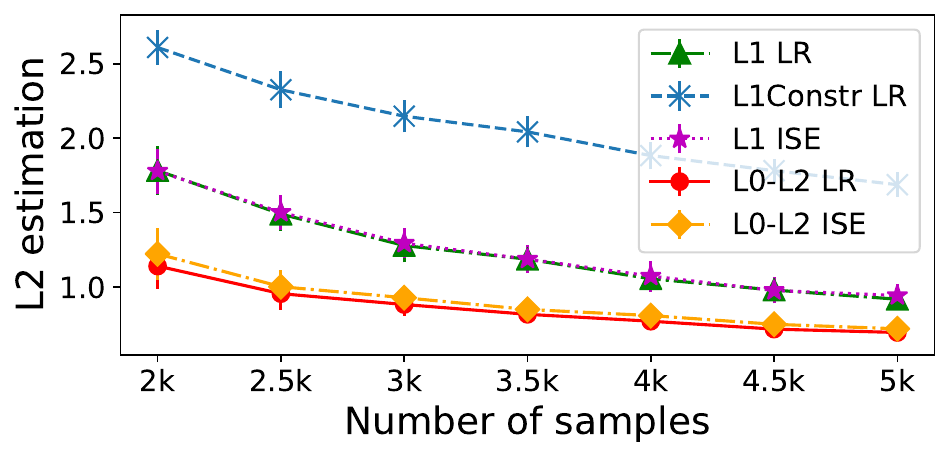}\label{fig:figb}
	\end{tabular}
	\smallskip
	\begin{tabular}{c c}
		{\sf{Example 1, $p=81$}} &  {\sf{Example 1, $p=100$}} \\
		\includegraphics[width=0.48\textwidth,height=0.18\textheight, clip = true ]{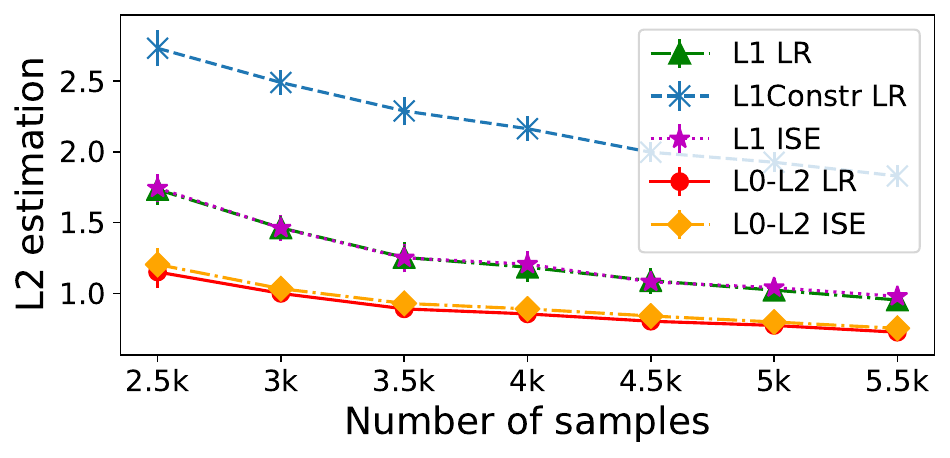}\label{fig:figa}&
		\includegraphics[width=0.48\textwidth,height=0.18\textheight, clip = true ]{periodic_graph_uniform_sign_P_100_rho_05_L2norm.pdf}\label{fig:figb}
	\end{tabular}
	%{\sf Example1}
	\caption{\small{ 
	Our methods additionally show important gains for L2 estimation. Note that, as predicted by theory, for the same number of samples, each method performance decreases when the graph size increases.
	}}
	\label{fig:first}
\end{figure*}

\section{Computational details}

\subsection{Proof of Proposition 1}
We first present the proof of the closed-form expression of the proximal operator. 
\newline
For $\bm{v} \in \mathbb{R}^{p-1}, k \in \mathbb{N}^*, \theta>0$, the proximal operator has been defined as
\begin{equation}\label{thresh-op-1}
\mathcal{S}(\bm{v} ; k; \theta ) := \argmin_{\B{w}\in \mathbb{R}^{p-1}: ~\|\B{w}\|_{0} \leq k, ~\|\B{w}\|_{2} \leq \theta}~~  \left\| \B{w} - \bm{v} \right\|_{2}^2.
\end{equation}
For a subset $\mathcal{S} \subset \mathbb{R}^{p-1}$ of size at most $k$: $|\mathcal{S}| \le k$, we consider the restriction of Problem \eqref{thresh-op-1} to $\mathcal{S}$:
\begin{equation}\label{thresh-op-2}
\argmin_{\B{w}\in \mathbb{R}^{p-1}: ~\|\B{w}\|_{2} \leq \theta}~~  \left\| \B{w} - \bm{v} \right\|_{2}^2
~~~ \sbt ~~~ w_i=0, ~ \forall i \notin \mathcal{S}.
\end{equation}
A solution of Problem \eqref{thresh-op-1} can be derived by first solving Problem \eqref{thresh-op-2} for any subset, then returning the solution with lower objective value.

\smallskip
\noindent
Let us fix  $\mathcal{S} \subset \mathbb{R}^{p-1}: |\mathcal{S}| \le k$. The objective value of Problem \eqref{thresh-op-2} can be expressed as:
\begin{equation}\label{thresh-op-3}
J(\mathcal{S}) = \min_{\B{w}_{\mathcal{S}} \in \mathbb{R}^{| \mathcal{S} |}: ~\|\B{w}_{\mathcal{S}}\|_{2} \leq \theta} ~~ \sum_{i \in \mathcal{S}} (w_i - v_i)^2 + \sum_{i \notin \mathcal{S}} v_i^2.
\end{equation}
We define $\B{v}_{\mathcal{S}} \in \mathbb{R}^{| \mathcal{S} |}$ as the restriction of $\B{v}$ into the indexes in $\mathcal{S}$. Hence, to solve Problem \eqref{thresh-op-2},  we define the projection of $\B{v}_{\mathcal{S}}$ onto the L2 ball of radius $\theta$:
$$ \hat{ \B{w}_{\mathcal{S}} } = \min\left(1,  \frac{\theta}{\| \B{v}_{\mathcal{S}} \|_2} \right) \B{v}_{\mathcal{S}} \in \mathbb{R}^{| \mathcal{S} |}.$$
We then extend $\hat{ \B{w}_{\mathcal{S}} }$ into a solution of Problem \eqref{thresh-op-2} by setting the coefficients outside $\mathcal{S}$ to $0$. Note that the Karush–Kuhn–Tucker conditions of Problem \eqref{thresh-op-2} can be used to assess the optimality of this extended solution. The objective value of Problem \eqref{thresh-op-2} becomes:
$$
J(\mathcal{S}) = \min(\| \B{v}_{\mathcal{S}} \|_2^2 - \theta^2, 0) + \sum_{i \notin \mathcal{S}} v_i^2.
$$
Therefore, to solve Problem \eqref{thresh-op-1}, we minimize the quantity $J(\mathcal{S})$ across all the subsets of size $k$. This minimization is achieved for the subset $\mathcal{S} = \{ (1), (2), \ldots, (k) \}$ of the $k$ largest entries of $\B{v}$.

\subsection{Proof of Proposition 2}
We then prove the convergence properties of the DFO algorithm presented in the main paper. Remember that, for $D\ge C$, for a non-negative differentiable loss $f$ with $C$-Lipschitz continuous gradient, it holds:
\begin{equation}\label{lipschitz-gradient}
f( \B{v} ) \le  Q_D(\B{u}, \B{v}) = f(\B{u}) + \nabla f(\B{u})^T(\B{v}-\B{u}) + \frac{D}{2} \| \B{v} - \B{u} \|_2^2, ~ \forall \B{u}, \B{v} \in \mathbb{R}^{p-1}.
\end{equation}
Consequently let us fix $\B{w} \in \mathbb{R}^{p-1}$ satisfying $\|\B{w}\|_{0} \le k$, $\|\B{w}\|_{2} \le \theta$ and, with the notations of the main paper, let us denote $\hat{\B{v}} \in \mathcal{S} \left(  \B{w} - \tfrac{1}{D} \nabla f( \B{w}) ; k; \theta \right)$. We then have:

\begin{align}
\begin{split}
f(\B{w})= Q_{D}(\B{w}, \B{w}) 
 &\ge \inf_{ \|\B{v} \|_{0} \le {k}, ~ \|\B{v} \|_{2} \le {\theta}  } \;\; Q_{D}(\B{w}, \B{v}) \\
 &= \inf_{ \|\B{v} \|_{0} \le {k}, ~ \|\B{v} \|_{2} \le {\theta}  } \;\; \left ( f(\B{w}) + \nabla f(\B{w})^T(\B{v}-\B{w}) + \frac{D}{2} \| \B{v} - \B{w} \|_2^2 \right )  \\
&=  \inf_{ \|\B{v} \|_{0} \le {k}, ~ \|\B{v} \|_{2} \le {\theta}  } \;\; \left (  \frac{D}{2} \left\| \B{v} - \left ( \B{w} - \frac{1}{D} \nabla f(\B{w}) \right) \right\|_2^2 - \frac{1}{2D} \|\nabla f(\B{w}) \|_2^2  + f(\B{w}) \right) \\
&=  f(\B{w}) + \nabla f(\B{w})^T(\hat{\B{v}} - \B{w}) + \frac{D}{2} \| \hat{\B{v}} - \B{w} \|_2^2 ~ \text{  as  } ~ \hat{\B{v}} \in \mathcal{S} \left(  \B{w} - \tfrac{1}{D} \nabla f( \B{w}) ; k; \theta \right) \\
&=f(\B{w}) + \nabla f(\B{w})^T(\hat{\B{v}} - \B{w}) + \frac{C}{2} \| \hat{\B{v}} - \B{w} \|_2^2 + \frac{D - C}{2} \| \hat{\B{v}} - \B{w} \|_2^2 \\
&= Q_{C}(\B{w}, \hat{\B{v}})  + \frac{D-C}{2} \| \hat{\B{v}} - \B{w}\|_2^2\\
&\ge f(\hat{\B{v}}) + \frac{D-C}{2} \| \hat{\B{v}} - \B{w} \|_2^2 ~ \text{ with Equation \eqref{lipschitz-gradient}.}
\end{split}
\end{align}
We consequently conclude that:
\begin{equation*}
f(\B{w}) - f(\hat{\B{v}}) \ge \frac{D-C}{2} \| \hat{\B{v}} - \B{w}\|_2^2.
\end{equation*}
In particular, when $\B{w} = \B{w}^{(t)}$, the DFO algorithm update gives $\hat{\B{v}} = \B{w}^{(t+1)}$ and we obtain:
\begin{equation}\label{suff-decrease-1}
f(\B{w}^{(t)}) - f(\B{w}^{(t+1)}) \ge \frac{D-C}{2} \| \B{w}^{(t+1)} - \B{w}^{(t)}\|_2^2.
\end{equation}
and we see that the sequence $\{f(\B{w}^{(t)}) \}_{t \ge 1}$ is decreasing. Because $f(\B{w}) \geq 0$, we conclude that the sequence $f(\B{w}^{(t)})$ converges to some $f^* \geq 0$.

\medskip

\noindent
Summing inequalities \eqref{suff-decrease-1} for $ 1 \leq t \leq T$, we obtain
\begin{equation}
\frac{D - C}{2}\sum_{t=1}^{T} \| \B{w}^{(t+1)} - \B{w}^{(t)}\|_2^2
\le \sum_{t=1}^{T} \left ( f ( \B{w}^{(t)}  ) - f( \B{w}^{(t+1)} )  \right ), 
\end{equation}
which leads to
$$ \frac{T (D - C) }{2}  \min_{1 \le t \le T}  \| \B{w}^{(t+1)} - \B{w}^{(t)}\|_2^2 \le f( \B{w}^{(1)} ) - f( \B{w}^{(t+1)} ).$$

\section{Statistical analysis}

\noindent
We consider $n$ independent realizations $\B{z}^{(1)}, \ldots, \B{z}^{(n)}$ from an Ising model with $p$ nodes, with connectivity matrix $\bm{W}^*$ and corresponding probability distribution:
\begin{equation}\label{ising}
p(\bm{z} | \bm{W}^*) = \frac{1}{Z(\bm{W}^*)} \exp\left( \frac{1}{2} \bm{z}^T \bm{W}^* \bm{z}  \right), ~\forall \bm{z} \in \{-1, 1\}^p.
\end{equation}
We introduce two subsets of $\mathbb{R}^{p-1}$ for our statistical analysis. First let us denote:
$$\mathcal{B}_{0, 2}(k, \lambda) = \{ \B{w} \in \mathbb{R}^{p-1}: ~ \| \B{w} \|_0 \le k,  ~ \| \B{w} \|_2 \le \lambda / \sqrt{k} \}.$$
We additionally introduce
$$\mathcal{B}_1(\lambda) = \{ \B{w} \in \mathbb{R}^{p-1}: ~ \| \B{w} \|_1 \le \lambda \},$$
and note the inclusion $\mathcal{B}_{0, 2}(k, \lambda) \subset \mathcal{B}_1(\lambda)$. 
%The existing literature define $\lambda^*$ such that $\B{w}^* \in \mathcal{B}_1(\lambda)$. 

\medskip

\noindent
For a node $j\in \{1, \ldots, p$, we have defined $\bm{w}^*_{-j} = (\bm{e}_j^T \bm{W}^*)_{-j} \in \mathbb{R}^{p-1}$ the $j$th row of the connectivity matrix without the diagonal term. We also have defined $(k^*, \lambda^*)$ as the infimum of the set of pairs such that $\B{w}^*_{-j} \in \mathcal{B}_2(k^*, \lambda^*)$ for all nodes.

\subsection{Statistical analysis for L0-L2 constrained logistic regression}

The L0-L2 constrained logistic regression procedure derives an estimate of the $j$th row of the connectivity matrix $\bm{w}^*_{-j}$ defined as
\begin{equation}\label{l0-l2-logreg}
\hat{\B w} ~~ \in \argmin \limits_{ \B{w} \in \mathcal{B}_{0, 2}(k, ~ \lambda) }  \mathcal{L}_n(\B{w})~~ \text{where}~~ \mathcal{L}_n(\B{w}) = \frac{1}{n} \sum_{i=1}^n \log \left(1 + \exp (- 2 y_i \B{x}_i^T  \B{w}) \right),
\end{equation}
where $k \ge k^*$, $\lambda \ge \lambda^*$ and we have defined  $y_i=z^{(i)}_j$ and $\bm{x}_i = \B{z}^{(i)}_{-j}$ as in the main paper. We drop the dependency upon $j$ when no confusion can be made. We prove herein the following Theorem 1.
\begin{theorem} \label{upper-bound-logreg}
Let $\delta \in (0,1)$. The L0-L2 constrained logistic regression estimator of the connectivity matrix  $\hat{\B{W}}_{\LR}$ derived by solving Problem \eqref{l0-l2-logreg} at each node for the parameters $k \ge k^*, \lambda \ge \lambda^*$  satisfies with probability at least $1 - \delta$:
\begin{equation*}
\| \hat{\B{W}}_{\LR} - \B{W}^* \|_{\infty}^2
\le  51 \lambda e^{4 \lambda} \sqrt{\frac{\log(e p(p-1) /k) }{n} \log(2 / \delta)}.
 \end{equation*}
\end{theorem}
We have defined the minimum absolute edge weight  $\eta$  of $\B{W}^*$ as $\eta = \min \left\{ |W^*_{ij}|: ~ W^*_{ij} \ne 0 \right\}.$
\medskip
\noindent
Hence, Theorem \ref{upper-bound-logreg} allows us to derive an upper bound on the number of samples required to recover with high probability the connectivity graph of the Ising model in Equation \eqref{l0-l2-logreg}.
\begin{cor}\label{cor1}
Let $\delta \in (0,1)$. Assume that $n$ satisfies $n \ge 204^2 \eta^{-4} \lambda^2 \exp(8 \lambda)\log\left( e p(p-1) / k \right) \log \left(2 / \delta \right)$. The L0-L2 constrained logistic regression estimator derived by hard-thresholding $\hat{\B{W}}_{\LR}$ with threshold $\eta /2$ recovers the exact connectivity graph with probability at least $1 -\delta$.
\end{cor}
\textbf{Proof: }
The proof immediately follows from selecting a number of samples $n$ such that:
\begin{equation}\label{necessary-recovery}
\| \hat{\B{W}}_{\LR} - \B{W}^* \|_{\infty} \le \frac{\eta}{2}, 
\end{equation}
in which case the triangle inequality guarantees that for any edge $(i,j)$ in the connectivity graph of $\B{W}^*$ it holds:
$$|\hat{W}_{\LR, ~ ij}| \ge | W^*_{ij} | - \frac{\eta}{2} \ge \frac{\eta}{4},$$
and the hard-thresholding procedure retains the edge $(i,j)$ in the graph estimate. Using Theorem \ref{upper-bound-logreg}, we conclude the proof by noting that Equation \eqref{necessary-recovery} is satisfied as soon as:
$$51 \lambda e^{4 \lambda} \sqrt{\frac{\log(e p(p-1) /k) }{n} \log(2 / \delta)} \le \frac{\eta^2}{4}.$$
\hfill$\square$

\smallskip
\noindent
We now present the proof of Theorem 1. Our proof is divided in four steps. We first prove the following Lemma \ref{first-lemma-logreg} in Section \ref{sec:first-lemma-logreg}.
\begin{lemma}\label{first-lemma-logreg}
Let $\delta \in (0, 1)$. We note $\tilde{\B{S}}_n = \frac{1}{n} \sum_{i=1}^n \B{x}_i \B{x}_i^T$. The empirical estimator $\hat{\B w}$ defined as a minimizer of Problem \eqref{l0-l2-logreg} at node $j$ satisfies with probability at least $1 - \delta / 2$:
\begin{equation}\label{lower-bound-logreg}
(\hat{\B{w}} - \B{w}^*)^T \tilde{\B{S}}_n (\hat{\B{w}} - \B{w}^*) 
= \frac{1}{n} \sum_{i=1}^n \left(\B{x}_i^T (\hat{\B{w}} - \B{w}^*) \right)^2 
\le 34 \lambda e^{2 \lambda} \sqrt{\frac{\log(e p(p-1) / k ) }{n} \log(2 / \delta)}.
\end{equation}
In addition, this relation holds simultaneously for all the $p$ empirical estimators of all the $p$ nodes of the Ising graph.
\end{lemma}

\noindent
Our second step derives a lower bound of the left-hand side of Equation \eqref{lower-bound-logreg}.
\begin{lemma}\label{supremum}
With the notations previously introduced, let us denote $\tilde{\B{S}} = \mathbb{E}(\tilde{\B{S}}_n)$. It then holds with probability at least $1 - \delta /2$:
$$
\sup_{\B{u} \in \mathcal{B}_{0, 2}(k, \lambda)} 
| \B{u}^T(\tilde{\B{S}}_n - \tilde{\B{S}} )\B{u} | \le 
17 \lambda^2 \sqrt{\frac{\log(e p/k)}{n} \log(2/ \delta}).
$$
In addition, this relation holds simultaneously for all the $p$ empirical estimators of all the $p$ nodes of the Ising graph.
\end{lemma}
Finally our last step proves the following result
\begin{lemma}\label{lower-bound} 
For $\B{u} \in \mathcal{B}_{0, 2}(k, \lambda)$ and $\tilde{\B{S}}$ defined as above, it holds:
$$
\B{u}^T \tilde{\B{S}} \B{u}
\ge \| \B{u} \|_{\infty}^2  e^{- 2\lambda}
$$
\end{lemma}
We prove Lemma \ref{supremum} in Section \ref{sec:supremum} and Lemma \ref{lower-bound} in Section \ref{sec:lower-bound} before pairing these results with Lemma \ref{first-lemma-logreg} to prove Theorem \ref{upper-bound-logreg} in Section \ref{sec:upper-bound-logreg}

\subsection{Statistical analysis for L0-L2 constrained interaction screening}

Our second estimator of $\bm{w}^*_{-j}$ is defined as a solution of the L0-L2 constrained interaction screening problem at node $j$:
\begin{equation}\label{l0-l2-ise}
\hat{\B w} ~~ \in \argmin \limits_{ \B{w} \in \mathcal{B}_{0, 2}(k, ~ \lambda) }  \mathcal{J}_n(\B{w})~~ \text{where}~~ \mathcal{J}_n(\B{w}) =  \frac{1}{n} \sum_{i=1}^n \exp (- y_i \B{x}_i^T  \B{w}).
\end{equation}
This estimator reaches a similar upper-bound, presented in the following theorem.
\begin{theorem} \label{upper-bound-ise}
Let $\delta \in (0,1)$. The L0-L2 constrained interaction screening estimator of the connectivity matrix  $\hat{\B{W}}_{\ISE}$ derived by solving Problem \eqref{l0-l2-ise} at each node for the parameters $k \ge k^*, \lambda \ge \lambda^*$  satisfies with probability at least $1 - \delta$:
\begin{equation*}
\| \hat{\B{W}}_{\ISE} - \B{W}^* \|_{\infty}^2
\le 153  (\lambda \vee \lambda^2) e^{4 \lambda} \sqrt{\frac{\log(e p(p-1) / k) }{n} \log(2 / \delta)}.
 \end{equation*}
\end{theorem}
Similar to Corollary \ref{cor1}, we derive the sample complexity for the L0-L2 constrained interaction screening procedure.
\begin{cor}
Let $\delta \in (0,1)$. Assume that $n \ge 612^2 \eta^{-4} (\lambda^2 \vee \lambda^4) \exp(8 \lambda)\log\left( e p(p-1) / k \right) \log \left(2 / \delta \right)$. The L0-L2 constrained logistic regression estimator derived by hard-thresholding $\hat{\B{W}}_{\LR}$ with threshold $\eta /2$ recovers the exact connectivity graph with probability at least $1 -\delta$.
\end{cor}
Theorem \ref{upper-bound-ise} is proved in Section \ref{sec:lemma-upper-bound-ise} and requires the same proof technique than for L0-L2 constrained logistic regression. More precisely, we pair Lemmas \ref{supremum} and \ref{lower-bound} with the following result, which is an adaptation of  Lemma \ref{first-lemma-logreg} to the L0-L2 constrained ISE.
\begin{lemma}\label{first-lemma-ise}
Let $\delta \in (0, 1)$. The empirical estimator $\hat{\B w}$ defined as a minimizer of Problem \eqref{l0-l2-ise} at node $j$ satisfies with probability at least $1 - \delta / 2$:
$$
\frac{1}{n} \sum_{i=1}^n \left(\B{x}_i^T (\hat{\B{w}} - \B{w}^*) \right)^2
\le 68\lambda (1 + \lambda) e^{2\lambda} \sqrt{\frac{\log(e p(p-1) / k ) }{n} \log(2 / \delta)}.
$$
In addition, this relation holds simultaneously for all the $p$ empirical estimators of all the $p$ nodes of the Ising graph.
\end{lemma}

\subsection{Preliminary results}
Our technical proof relies on supremum analysis of sub-Gaussian random variables. First, we recall here the definition of a sub-Gaussian random variable~\citep{lecture-notes}:
\begin{defn} \label{def-asu3}
	A random variable $Z$ is said to be sub-Gaussian with variance $\sigma^2>0$ if $\mathbb{E}(Z) = 0$ and $\mathbb{E}( \exp(t Z)) \le \exp \left( \frac{\sigma^2 t^2}{2} \right), \  \forall t>0$. 
\end{defn}
This variable will be noted $Z \sim \subg(\sigma^2)$. Let us additionally introduce an additional lemma which controls the supremum of sub-Gaussian random variables. The lemma is an extension for sub-Gaussian random variables of Proposition E.1, \cite{bellec2018slope}.

\begin{lemma}\label{upper-bound-sup} 
    (Lemma 5, \cite{dedieu2019error})
    Let $g_1,\ldots g_r$ be sub-Gaussian random variables with variance $\sigma^2$. We denote by $(g_{(1)}, \ldots, g_{(r)})$ a non-increasing rearrangement of $(|g_1|, \ldots, |g_r|)$ and define the coefficients $\lambda_j^{(r)} = \sqrt{ \log(2r/j) }, \ j=1,\ldots\,r$. For $\delta \in \left(0, \frac{1}{2} \right)$, it holds with probability at least $1-\delta$:
    $$ \sup \limits_{j=1,\ldots,r} \left\{ \frac{ g_{(j)} }{\sigma \lambda_j^{(r)}} \right\} \le 12 \sqrt{ \log(1 / \delta)}.  $$ 
\end{lemma}
(Lemma 5 in \cite{dedieu2019error} wrongly states that the variable have to be independent: this assumption is not used in the proof presented. Also see Lemma 4, \cite{dedieu2019sparse})

\subsection{Three useful events for independent realizations from an Ising model}
The following Lemmas \ref{event-A}, \ref{event-B} and \ref{event-C} introduce three useful events applying to the observations $\B{z}^{(1)}, \ldots, \B{z}^{(n)}$ that we later use in our analysis.
\begin{lemma}\label{event-A}
Let us denote $g_{j \ell} = \frac{1}{\sqrt{n}} \sum \limits_{i=1}^n \left( z^{(i)}_j - \mathbb{E}\left(z^{(i)}_j |\B{z}^{(i)}_{-j} \right) \right) z^{(i)}_{\ell}, \forall j, ~\ell \ne j$. 
We define $\B{H} \in \mathbb{R}^{p(p-1)}$  such that $H_{(p-1)j + \ell - \mathbf{1}_{\ell > j} } = g_{j \ell}$ and we introduce a non-increasing rearrangement $(H_{(1)}, \ldots, H_{(p(p-1))})$  of $(|H_1|, \ldots, |H_{p(p-1)}|)$. With the notations of Lemma \ref{upper-bound-sup}, we define the event:
\begin{equation*}
\mathcal{A} = \left\{ \sup_{\ell=1,\ldots, p(p-1)} \frac{H_{(\ell) }}{\lambda_{\ell}^{(p(p-1))} } \le 12 \sqrt{\log(2 / \delta)} \right\}.
\end{equation*}
It then holds $$p(\mathcal{A}) \ge 1 - \delta / 2.$$
\end{lemma}
Note: The indicator $\mathbf{1}_{\ell > j}$ is only used for proper indexing.

\begin{Proof}
Following Hoeffding's theorem (Theorem 1.9, \cite{lecture-notes}), %any centered random variable almost surely bounded by $M>0$ is sub-Gaussian with variance $M^2$. Consequently,
because $\|\B{z}^{(i)}\|_{\infty} \le 1, \forall i$, it holds:
$$\left( z^{(i)}_j - \mathbb{E}\left(z^{(i)}_j |\B{z}^{(i)}_{-j} \right) \right) z^{(i)}_{\ell} \sim \subg(1), ~\forall i,~j,~\ell \ne j,$$
where the law of total  expectation guarantees that $\mathbb{E}\left(z^{(i)}_j z^{(i)}_{\ell}\right) = 
\mathbb{E}\left( \mathbb{E}\left(z^{(i)}_j |\B{z}^{(i)}_{-j} \right) z^{(i)}_{\ell} \right)$ and the variables $g_{j \ell}$ have zero mean.

\noindent
In addition, the sub-Gaussian random variables $\left( z^{(i)}_j - \mathbb{E}\left(z^{(i)}_j |\B{z}^{(i)}_{-j} \right) \right) z^{(i)}_{\ell} , ~ i=1,\ldots,n$ are independent with variance $1$. We consequently know that:
$$g_{j \ell} = \frac{1}{\sqrt{n}} \sum \limits_{i=1}^n \left( z^{(i)}_j - \mathbb{E}\left(z^{(i)}_j |\B{z}^{(i)}_{-j} \right) \right) z^{(i)}_{\ell} \sim \subg(1).$$
Lemma \ref{event-A} follows from applying Lemma \ref{upper-bound-sup} to the $p^2 - p$ sub-Gaussian random variables $g_{j \ell}$.
\end{Proof}

\medskip

\begin{lemma}\label{event-B}
Let us denote the empirical matrix $\B{S}_n = \frac{1}{n} \sum \limits_{i=1}^n \B{z}^{(i)} \B{z}^{(i)~T} \in \mathbb{R}^{p^2}$ and $\B{S} = \mathbb{E}(\B{S}_n)$ its theoretical counterpart. We define $\B{T} \in \mathbb{R}^{p^2}$  such that $T_{p\ell + m} = (\B{S}_n - \B{S})_{\ell m}$ and we introduce a non-increasing rearrangement $(T_{(1)}, \ldots, T_{(p^2)})$  of $(|T_1|, \ldots, |T_{p^2}|)$. We define the event:
\begin{equation*}
\mathcal{B} = \left\{ \sup_{\ell=1,\ldots,p^2} \frac{T_{(\ell)}}{\lambda_{\ell}^{(p^2)} } \le 12 \sqrt{ \frac{\log(2 / \delta)}{n} } \right\}.
\end{equation*}
It then holds $$p(\mathcal{B}) \ge 1 - \delta / 2.$$
\end{lemma}

\begin{Proof}
Let us first note that the entry with indices  $(\ell,m)$ of $(\B{S}_n - \B{S})$ is equal to:
$$(\B{S}_n - \B{S})_{\ell m} = \frac{1}{n} \sum_{i=1}^n \left(z^{(i)}_{\ell} z^{(i)}_m - \mathbb{E}\left(z^{(i)}_{\ell} z^{(i)}_m \right) \right), ~ \forall \ell,m.$$
Because the above variables have zero mean, the observations are independent and the entries are bounded by 1, Hoeffding's theorem gives us:
$$(\B{S}_n - \B{S})_{\ell m} \sim \subg \left(\frac{1}{n} \right).$$
%We proceed similarly to Equation \eqref{lemma1-second-part}.  We define $\B{T}, \B{v} \in \mathbb{R}^{(p-1)^2}$  such that $T_{i * p + j} = (\B{S}_n - \B{S})_{ij}.$
Similarly than above, Lemma \ref{event-B} follows from applying Lemma \ref{upper-bound-sup} to the $p^2$ sub-Gaussian random variables $(\B{S}_n - \B{S})_{\ell m}$.
\end{Proof}

\begin{lemma}\label{event-C}
Finally, let us denote $\beta_{j \ell} = \frac{1}{\sqrt{n}} \sum \limits_{i=1}^n z^{(i)}_{j} \exp \left(- z^{(i)}_{j} (\B{z}^{(i)}_{-j})^T  \B{w}^* \right) z^{(i)}_{\ell} ~ \forall j, ~\ell \ne j$. We define $\B{\Gamma} \in \mathbb{R}^{p(p-1)}$  such that $\Gamma_{(p-1)j + \ell - \mathbf{1}_{\ell > j}} = \beta_{j \ell}$ and we introduce a non-increasing rearrangement $(\Gamma_{(1)}, \ldots, \Gamma_{(p(p-1))})$  of $(|\Gamma_1|, \ldots, |\Gamma_{p(p-1)}|)$. With the notations of Lemma \ref{upper-bound-sup}, we define the event:
\begin{equation*}
\mathcal{C} = \left\{ \sup_{\ell=1,\ldots, p(p-1)} \frac{\Gamma_{(\ell) }}{\lambda_{\ell}^{(p(p-1))} } \le 12 e^{\lambda} \sqrt{\log(2 / \delta)} \right\}.
\end{equation*}
It then holds $$p(\mathcal{C}) \ge 1 - \delta / 2.$$
\end{lemma}
\begin{Proof}
First, let us observe that the variables $\beta_{j \ell}$ have zero mean as it holds:
\begin{align*}
\begin{split}
\mathbb{E}(\beta_{j \ell})   
&= \mathbb{E}\left( \mathbb{E}\left(\beta_{j \ell} | \B{z}^{(i)}_{-j} \right) \right)
= \mathbb{E}\left( \frac{1}{\sqrt{n}} \sum \limits_{i=1}^n  z^{(i)}_{\ell}
\mathbb{E}\left( z^{(i)}_{j} \exp \left(- z^{(i)}_{j} (\B{z}^{(i)}_{-j})^T  \B{w}^* \right) \bigg| \B{z}^{(i)}_{-j} \right) \right)\\
&=\mathbb{E}\left( \frac{1}{\sqrt{n}} \sum \limits_{i=1}^n  z^{(i)}_{\ell} \left\{ \frac{\exp \left(- (\B{z}^{(i)}_{-j})^T  \B{w}^* \right)}{1 + \exp \left(- 2 (\B{z}^{(i)}_{-j})^T  \B{w}^* \right)}
-  \frac{\exp \left((\B{z}^{(i)}_{-j})^T  \B{w}^* \right)}{1 + \exp \left( 2 (\B{z}^{(i)}_{-j})^T  \B{w}^* \right)} \right\}
\right) = 0.\\
\end{split}
\end{align*}
In addition, because $\|\B{z}^{(i)}\|_{\infty} \le 1, \forall i$ and $\| \B{w}^* \|_{1} \le \lambda$, it holds:
$$\left| z^{(i)}_{j} \exp \left(- z^{(i)}_{j} (\B{z}^{(i)}_{-j})^T  \B{w}^* \right) z^{(i)}_{\ell} \right| \le e^{\lambda}$$
Similarly to the above, we conclude the proof by applying Lemma \ref{upper-bound-sup} to the $p^2-p$ sub-Gaussian random variables
$$\beta_{j \ell} \sim \subg \left(e^{2\lambda} \right),  ~ \forall j, \ell \ne j.$$
\end{Proof}

\subsection{Proof of Lemma \ref{first-lemma-logreg}}\label{sec:first-lemma-logreg}

\noindent
We fix $j \in \{1,\ldots,p\}$ and drop the dependency upon $j$. In particular, we note $\B{w}^* = \B{w}^*_{-j}$. We recall that the empirical estimator $\hat{\B w}$ at node $j$ is defined as a minimizer of Problem \eqref{l0-l2-logreg}. 

\medskip
\noindent
Our approach follows Lemma 5.21, \cite{lecture-notes}, and leverage the additional information that  $\hat{\B w}$ and $\B{w}^*$ are $k$-sparse vectors with bounded L2 norm. 
First, let us note that:
\begin{align*}
\log \left(1 + \exp (- 2 y_i \B{x}_i^T  \B{w})\right)
&= \log \left(\frac{\exp (y_i \B{x}_i^T  \B{w}) + \exp (- y_i \B{x}_i^T  \B{w})}{\exp (y_i \B{x}_i^T  \B{w})} \right)\\
&= \log \left(\frac{\exp (\B{x}_i^T  \B{w}) + \exp (- \B{x}_i^T  \B{w}) }{\exp ((1 + y_i) \B{x}_i^T  \B{w})
\exp (- \B{x}_i^T  \B{w}) }\right) \\
&= \log \left(1 + \exp (2 \B{x}_i^T  \B{w})\right) 
- (1 + y_i) \B{x}_i^T  \B{w}.
\end{align*}
The empirical logistic loss can then be expressed as:
$$\mathcal{L}_n({\B w}) 
= \frac{1}{n} \sum_{i=1}^n \left\{ \log \left(1 + \exp (2 \B{x}_i^T  \B{w})\right) - (1 + y_i) \B{x}_i^T  \B{w} \right\}.$$
We also define the theoretical logistic loss as:
$$\mathcal{L}({\B w}) 
= \mathbb{E}\left( \mathcal{L}_n({\B w}) | \B{x}_i \right)
= \frac{1}{n} \sum_{i=1}^n \left\{ \log \left(1 + \exp (2 \B{x}_i^T  \B{w})\right) - (1 + \mathbb{E}(y_i | \B{x}_i) ) \B{x}_i^T  \B{w} \right\},$$
$\mathcal{L}$ is convex and its gradient is
\begin{equation}\label{grad}
\nabla \mathcal{L}(\B{w}) 
=  \frac{1}{n} \sum_{i=1}^n
\frac{2}{1 + \exp ( - 2 \B{x}_i^T  \B{w})} \B{x}_i
-  (1 + \mathbb{E}(y_i | \B{x}_i)) \B{x}_i.
\end{equation}
In addition, let us note that:
$$\mathbb{E}(y_i | \B{x}_i) = \frac{\exp (\B{x}_i^T  \B{w}^* ) - \exp ( - \B{x}_i^T  \B{w}^* )}{\exp (\B{x}_i^T  \B{w}^* ) + \exp ( - \B{x}_i^T  \B{w}^* )},$$
which we can plug into Equation \eqref{grad} to conclude that $\nabla \mathcal{L}(\B{w}^*)=0$ and that $\B{w}^*$ minimizes the (convex) theoretical logistic loss $\mathcal{L}$. Because $\B{w}^*$ satisfies $\| \B{w}^* \|_0 \le k$, $\| \B{w}^*  \|_2 \le \lambda / \sqrt{k}$, we then have:
\begin{equation} \label{def-beta0}
\B{w}^* ~~ \in~~ \argmin \limits_{ \B{w} \in \mathcal{B}_{0, 2}(k, \lambda) }  \mathcal{L}(\B{w}),
\end{equation}
which is the theoretical counterpart of Problem \eqref{l0-l2-logreg}. Let us denote $\B{h}=\hat{\B{w}} - \B{w}^* $. Because $\hat{\B w}$ is a minimizer of the empirical loss, then $\mathcal{L}_n(\hat{\B w}) \le \mathcal{L}_n(\B{w}^* )$ and we consequently have:
\begin{align}\label{lemma1-first-part}
\begin{split}
\mathcal{L}(\hat{\B w}) - \mathcal{L}(\B{w}^* )
&\le \mathcal{L}(\hat{\B w}) - \mathcal{L}(\B{w}^* )
+ \mathcal{L}_n(\B{w}^* ) - \mathcal{L}_n(\hat{\B w})\\
&= \left( \mathcal{L}(\hat{\B w} ) - \mathcal{L}_n(\hat{\B w} ) \right) - \left( \mathcal{L}(\B{w}^* ) - \mathcal{L}_n(\B{w}^* ) \right) \\
&= \frac{1}{n} \sum_{i=1}^n (y_i - \mathbb{E}(y_i | \B{x}_i)) \B{x}_i^T \hat{\B w}- \frac{1}{n} \sum_{i=1}^n (y_i - \mathbb{E}(y_i | \B{x}_i)) \B{x}_i^T \B{w}^*\\
&= \frac{1}{n} \sum_{i=1}^n (y_i - \mathbb{E}(y_i | \B{x}_i)) \B{x}_i^T \left( \hat{\B w} - \B{w}^* \right)\\
&= \frac{1}{\sqrt{n}} \sum_{\ell=1}^{p-1} \left\{ \frac{1}{\sqrt{n}} \sum_{i=1}^n (y_i - \mathbb{E}(y_i | \B{x}_i)) x_{i\ell} \right\} h_{\ell}.
\end{split}
\end{align}
To upper-bound the quantity $ \mathcal{L}(\B{w}^* ) - \mathcal{L}(\hat{\B w})$ in  Equation \eqref{lemma1-first-part}, let us define the random variables $\tilde{g}_{\ell} = \frac{1}{\sqrt{n}} \sum_{i=1}^n (y_i - \mathbb{E}(y_i | \B{x}_i)) x_{i \ell}, \forall \ell \in \left\{1, \ldots, p-1\right\}$ and observe that $\tilde{g}_{\ell}=g_{j \ell}=H_{(p-1)j + \ell}$ where the  random variables $g_{j \ell}, ~ H_{(p-1)j + \ell}$ have been defined in Lemma \ref{event-A}. 
\medskip

\noindent
We consequently assume that the event $\mathcal{A}$ is satisfied, and also assume without loss of generality that $|h_1| \ge \ldots \ge |h_{2k}| \ge |h_{2k + 1}| = \ldots = |h_p|=0$. Lemma \ref{event-A} gives, with probability at least $1-\frac{\delta}{2}$:

\begin{align}\label{lemma1-second-part}
\begin{split}
\mathcal{L}(\hat{\B w}) - \mathcal{L}(\B{w}^* )
&\le \frac{1}{\sqrt{n}} \sum_{\ell=1}^{p-1} | \tilde{g}_{\ell} | |  h_{\ell} |
= \frac{1}{\sqrt{n}} \sum_{\ell=1}^{p-1} | H_{(p-1)j + \ell} | ~ |  h_{\ell} |\\ 
&\le \frac{1}{\sqrt{n}} \sum_{\ell=1}^{p-1} H_{(\ell)} | h_{\ell} | \text{ since } H_{(1)} \ge \ldots \ge H_{(p(p-1))}  \text { and }  | h_1 | \ge \ldots \ge |h_p|\\
&\le  \frac{1}{\sqrt{n}} \sum_{\ell=1}^{p-1} \frac{H_{(\ell)}}{ \lambda_{\ell}^{(p(p-1))} } \lambda_{\ell}^{(p(p-1))} |h_{\ell}| \text{ with the notations of Lemma } \ref{event-A}\\
&\le \frac{1}{\sqrt{n}} \sup_{\ell=1,\ldots,p(p-1)} \left\{\frac{H_{(\ell)}}{ \lambda_{\ell} }  \right\}   \sum_{\ell=1}^{p-1} \lambda_{\ell}  |h_{\ell} | \text{ where we have noted } \lambda_{\ell} =\lambda_{\ell}^{(p(p-1))} \\
&\le 12 \sqrt{\frac{\log(2/ \delta)}{n}} \sum_{\ell=1}^p \lambda_{\ell}  |h_{\ell} |\text{ with Lemma \ref{event-A} since }\mathcal{A} \text{ is satisified.}\\
&\le 12 \sqrt{\frac{\log(2/ \delta)}{n}} \sum_{\ell=1}^{2k} \lambda_{\ell}  |h_{\ell} | \text{ since } |h_{2k + 1}| = \ldots = |h_{p}| =0.
\end{split}
\end{align}
Cauchy-Schwartz inequality leads to:
\begin{align*}
\sum_{\ell=1}^{2k} \lambda_{\ell} | h_{\ell} |  &\le \sqrt{\sum_{\ell=1}^{2k} \lambda_{\ell} ^2 } \| \B{h}  \|_2 \le \sqrt{2k\log(2p(p-1)e /2k)}  \| \B{h}  \|_2 \le 2\lambda \sqrt{2\log(e p(p-1) /k)} 
\end{align*}
where we have used that $\| \B{h}  \|_2 \le 2 \lambda / \sqrt{k}$ and the Stirling formula to obtain
\begin{align*}
\sum_{\ell=1}^{2k} \lambda_{\ell} ^2 = \sum_{\ell=1}^{2k}  \log(2p(p-1)/ \ell) = 2k  \log(2p(p-1)) - \log((2k) !) 
&\le 2k \log(2p(p-1)) - 2k\log(2k/ e) \\
&= 2k \log(2e p(p-1) / 2k).
\end{align*}
Equation \eqref{lemma1-second-part} consequently gives, with probability at least $1- \delta / 2$:
\begin{equation}\label{lemma1-first-half}
\mathcal{L}(\hat{\B w}) - \mathcal{L}(\B{w}^* )
\le 34 \lambda \sqrt{\frac{\log(e p(p-1) / k ) }{n} \log(2 / \delta)}.
\end{equation}
We now lower-bound the left-hand size of Equation \eqref{lemma1-first-half}. Because $\nabla \mathcal{L}(\B{w}^*)=0$, a Taylor formula around $\B{w}^* $ gives us:
\begin{equation}\label{taylor}
\mathcal{L}(\hat{\B w}) - \mathcal{L}(\B{w}^* ) = 
(\hat{\B{w}} - \B{w}^*)^T \nabla^2 \mathcal{L}(\B{v}) (\hat{\B{w}} - \B{w}^*),
\text{where} ~ \B{v} = t \B{w}^* + (1-t) \hat{\B w} ~
\text{for some} ~ t \in (0,1).
\end{equation}
It holds that $\| \B{v} \|_1 \le t \| \B{w}^* \|_1 + (1-t) \| \hat{\B w} \|_1 \le \lambda$. In addition, Following Equation \eqref{grad}, the Hessian of $\mathcal{L}$ evaluated at $\B{v}$ is:
$$
\nabla^2 \mathcal{L}({\B v}) 
= \frac{1}{n} \sum_{i=1}^n  \frac{4\exp ( - 2 \B{x}_i^T  \B{v})}{ \left(1 + \exp ( - 2 \B{x}_i^T  \B{v} )\right)^2 } \B{x}_i \B{x}_i^T.
$$
Because $\| \B{x}_i \|_{\infty} \le 1$ and $\| \B{v} \|_1 \le \lambda$, it holds $| \B{x}_i^T  \B{v}| \le \lambda$. We then have:
\begin{align}\label{lower-bound-hessian}
\begin{split}
\frac{4\exp (- 2 \B{x}_i^T  \B{v} )}{ \left(1 + \exp ( -2 \B{x}_i^T  \B{v} )\right)^2 } 
&= \frac{4}{(1 + \exp (-2 \B{x}_i^T  \B{v} )) (1 + \exp ( 2 \B{x}_i^T  \B{v} ))}\\
&= \frac{4}{2 + \exp (-2 \B{x}_i^T  \B{v} ) + \exp ( 2 \B{x}_i^T  \B{v} )}\\
&\ge \frac{2}{1 + \exp (2 | \B{x}_i^T  \B{v} | )}
\ge \frac{2}{1 + \exp (2 \lambda )}
\ge \frac{2}{2 \exp (2 \lambda )} = e^{-2\lambda}
\end{split}
\end{align}
Pairing Equations \eqref{lemma1-first-half}, \eqref{taylor} and \eqref{lower-bound-hessian} we conclude that, with probability at least $1- \delta / 2$:
\begin{equation}\label{conclusion-lemma1}
\frac{1}{n} \sum_{i=1}^n \left(\B{x}_i^T (\hat{\B{w}} - \B{w}^*) \right)^2 \le 34 \lambda e^{2 \lambda} \sqrt{\frac{\log(e p (p-1) / k ) }{n} \log(2 / \delta)}.
\end{equation}
This relation holds simultaneously for the $p$ minimizers for all the nodes of the Ising graph, as the event $\mathcal{A}$ is shared across all the $p$ nodes.

\subsection{Proof of Lemma \ref{supremum}}\label{sec:supremum}

Let $\B{u} \in \mathcal{B}_{0, 2}(k, \lambda)$. 
\newline
Our first step to prove Lemma \ref{supremum} is to observe that the matrix $\tilde{\B{S}}_n$ is a submatrix of size $(p-1) \times (p-1)$ of the matrix $\B{S}_n$ introduced in Lemma \ref{event-B}. We consequently assume that the event $\mathcal{B}$ defined in Lemma \ref{event-B} is satisfied. We define $\B{T} \in \mathbb{R}^{p^2}$ as in Lemma \ref{event-B}, and denote $\B{v} \in \mathbb{R}^{(p-1)^2}$ such that $v_{(p -1)\ell + m} = u_{\ell} u_m$. We assume without loss of generality that $|v_1| \ge \ldots \ge |v_{k^2}| \ge |v_{k^2 + 1}| = \ldots = |v_{(p-1)^2}| = 0$, where we have used that $\| \B{u} \|_0 = k$.  Then, it holds with probability at most $1 - \delta / 2$:
\begin{align*}
\begin{split}
| \B{u}^T(\tilde{\B{S}}_n - \tilde{\B{S}})\B{u} |
&\le \frac{1}{n} \sum_{1 \le \ell, m \le p-1} | (\tilde{\B{S}}_n - \tilde{\B{S}})_{\ell m} |
| u_{\ell} u_m |\\
&\le \frac{1}{n} \sum_{\ell=1}^{(p-1)^2} | T_{(\ell)} | | v_{\ell} | \text{ since } T_{(1)} \ge \ldots \ge T_{(p^2)} \text { and }  | v_1 | \ge \ldots \ge |v_{(p-1)^2}| \\
&\le \sup_{\ell=1,\ldots,p^2} \left\{\frac{T_{(\ell)}}{\lambda_{\ell}^{(p^2)} }  \right\} ~~~  \sum_{\ell=1}^{(p-1)^2} \lambda_{\ell}^{(p^2)}  |v_{(\ell)} | \\
&\le 12 \sqrt{\frac{\log(2/ \delta)}{n}} \sum_{\ell=1}^{(p-1)^2} \lambda_{\ell}  |v_{(\ell)} |\text{ with Lemma \ref{event-B} and by noting } \lambda_{\ell}=\lambda_{\ell}^{(p^2)}\\
&\le 12 \sqrt{\frac{\log(2/ \delta)}{n}} \left( \sum_{\ell=1}^{k^2} \lambda_{\ell}  |v_{\ell} |  \right) \text{ since }  v_{k^2 + 1} = \ldots = v_{(p-1)^2} = 0\\
&\le 12 \sqrt{\frac{\log(2/ \delta)}{n}}\sqrt{k^2 \log(2e p^2 /k^2)} \| \B{v} \|_2\\
&\le 12 \sqrt{\frac{\log(2/ \delta)}{n}}\sqrt{2 k^2 \log(e p /k)} \| \B{v} \|_2 \text { as } 2 \le e
\end{split}
\end{align*}
Note that this results holds uniformly over $\mathcal{B}_{0, 2}(k, \lambda)$ as the event $\mathcal{B}$ considered does not depend upon $\B{u}, \B{v}$. 
In addition, let us note that:
$$\| \B{v} \|_2^2
= \sum_{1 \le \ell,m \le p-1} u_{\ell}^2 u_m^2
= \| \B{u} \|_2^4
\le \lambda^4 /k^2.
$$
We then conclude that it holds with probability at least $1 - \delta/2$:
\begin{equation}
\sup_{\B{u} \in \mathcal{B}_{0, 2}(k, \lambda)} | \B{u}^T(\tilde{\B{S}}_n - \tilde{\B{S}})\B{u} | \le 
17 \lambda^2 \sqrt{\frac{\log(ep /k)}{n} \log(2/ \delta}).
\end{equation}
This relation holds for the $p$ minimizers for all the nodes of the Ising graph, as the event $\mathcal{B}$ is shared.

\subsection{Proof of Lemma \ref{lower-bound}}\label{sec:lower-bound}

Let $\B{z}=(\B{x}, y)$ be a realization of the Ising model defined in Equation \eqref{ising}, where $y=z_j$ as previously. %Let us note that $p(\B{x} | \bm{W}^*) = p( - \B{x} | \bm{W}^*)$, and consequently $\mathbb{E}( \B{x}^T  \B{u}) = 0$. We then derive:
%$$\Var(\B{x}^T  \B{u})
%= \mathbb{E}( (\B{x}^T  \B{u})^2 )
%= \B{u}^T \tilde{\B{S}} \B{u}.$$
Let $\B{u} \in \mathcal{B}_{0, 2}(k, \lambda)$ and let us assume without loss of generality that $u_1 = \| \B{u} \|_{\infty}$. We then have:
\begin{align}\label{lemma3-first-part}
\begin{split}
\B{u}^T \tilde{\B{S}} \B{u}
&= \mathbb{E}( (\B{x}^T  \B{u})^2 )
= \mathbb{E} \left( \left( x_1 u_1 + \sum_{j=2}^{p-1} x_j u_j
\right)^2 \right)\\
&= u_1^2 + \mathbb{E} \left( \left(\sum_{j=2}^{p-1} x_j u_j
\right)^2 \right) + 2\mathbb{E} \left( x_1 u_1 \left(\sum_{j=2}^{p-1} x_j u_j \right) \right) 
\end{split}
\end{align}
Because $2|ab| \le a^2 + b^2, \forall a, b$ it holds with the law of total expectation:
\begin{align}\label{lemma3-second-part}
\begin{split}
2\mathbb{E} \left( x_1 u_1 \left(\sum_{j=2}^{p-1} x_j u_j \right) \right) 
&= 2\mathbb{E} \left( \mathbb{E} \left( x_1 u_1 \left(\sum_{j=2}^{p-1} x_j u_j \right) \bigg| \B{x}_{-1} \right)\right)
= 2\mathbb{E} \left( u_1 \mathbb{E}(x_1 | \B{x}_{-1}) \left(\sum_{j=2}^{p-1} x_j u_j \right) \right)\\
&\le u_1^2 \mathbb{E} \left( \mathbb{E} \left(x_1 | \B{x}_{-1} \right)^2 \right)
+ \mathbb{E} \left( \left(\sum_{j=2}^{p-1} x_j u_j \right)^2 \right),
\end{split}
\end{align}
where $\B{x}_{-1}=(x_2, \ldots, x_{p-1}) \in \mathbb{R}^{p-2}$.

\smallskip
\noindent
By using properties of the conditional expectation and Jensen inequality:
\begin{align}\label{lemma3-third-part}
\begin{split}
\mathbb{E} \left\{ \mathbb{E} \left(x_1 | \B{x}_{-1} \right)^2 \right\}
&= \mathbb{E} \left\{  \mathbb{E} \left( \mathbb{E} \left(x_1 | \B{x}_{-1}, y \right) | \B{x}_{-1} \right)^2 \right\}\\
&\le \mathbb{E} \left\{ \mathbb{E} \left( \mathbb{E} (x_1 | \B{x}_{-1}, y )^2 | \B{x}_{-1} \right) \right\}= \mathbb{E} \left\{ \mathbb{E} (x_1 | \B{x}_{-1}, y )^2 \right\}\\
&\le \sup_{\xi \in \{-1, 1\}^{p-1} } \mathbb{E} (x_1 | \left(\B{x}_{-1}, y\right) = \B{\xi} )^2\\
&=\left(\sup_{\B{\xi} \in \{-1, 1\}^{p-1} }  \frac{\exp(\B{\xi}^T \B{w}^*_{-1}) - \exp(- \B{\xi}^T \B{w}^*_{-1})}{\exp(\B{\xi}^T \B{w}^*_{-1}) + \exp(- \B{\xi}^T \B{w}^*_{-1})} \right)^2\\
&= \left(\sup_{\B{\xi} \in \{-1, 1\}^{p-1} }  \frac{\exp(2 \B{\xi}^T \B{w}^*_{-1}) - 1}{\exp(2\B{\xi}^T \B{w}^*_{-1}) + 1} \right)^2
\end{split}
\end{align}
Because $\psi: t \mapsto \frac{e^{2t} - 1}{e^{2t} + 1}$ is decreasing over $\mathbb{R}^-$ and increasing over $\mathbb{R}^+$, the supremum of Equation \eqref{lemma3-third-part} is reached for $\B{\xi} = \pm \sign(\B{w}^*_{-1})$ (where the $\sign$ function is applied componentwise) for which $|\B{\xi}^T \B{w}^*_{-1}| = \| \B{w}^*_{-1} \|_1 \le \lambda $. Equation \eqref{lemma3-third-part} consequently gives:
\begin{equation}\label{lemma3-fourth-part}
\mathbb{E} \left\{  \mathbb{E} \left(x_1 | \B{x}_{-1} \right)^2 \right\} \le \left(\frac{e^{2 \lambda} - 1}{e^{2\lambda} + 1} \right)^2.
\end{equation}
By pairing Equations \eqref{lemma3-first-part}, \eqref{lemma3-second-part} and \eqref{lemma3-fourth-part}, we conclude that:
\begin{align*}
\begin{split}
\B{u}^T \tilde{\B{S}}  \B{u}
\ge \| \B{u} \|_{\infty}^2 \left(1 - \left(\frac{e^{2 \lambda} - 1}{e^{2\lambda} + 1} \right)^2 \right)
&= \| \B{u} \|_{\infty}^2 \frac{2e^{2 \lambda}}{e^{2\lambda} + 1} \frac{2}{e^{2\lambda} + 1} 
\ge \| \B{u} \|_{\infty}^2  e^{- 2\lambda}.
\end{split}
\end{align*}

\paragraph{Remark: } Lemma \ref{lower-bound} holds more generally over $\mathcal{B}_1(\lambda)$ as we do not exploit the sparsity of $\B{u}$.

\subsection{Proof of Theorem \ref{upper-bound-logreg}}\label{sec:upper-bound-logreg}
We can now prove our upper bound for the L0-L2 constrained logistic regression estimator defined as a solution of Problem \eqref{l0-l2-logreg}. Pairing the results of Lemmas \ref{supremum} and \ref{lower-bound}, we obtain uniformly on ${B}_{0, 2}(k, \lambda)$, with probability at least $1 - \delta /2$:
\begin{align*}
\begin{split}
\B{u}^T \B{S}_n \B{u} 
&\ge \B{u}^T \B{S} \B{u} - 17 \lambda^2 \sqrt{\frac{\log(ep /k)}{n} \log(2/ \delta})\\
&\ge \| \B{u} \|_{\infty}^2  e^{- 2\lambda} - 17 \lambda^2 \sqrt{\frac{\log(ep /k)}{n} \log(2/ \delta})
\end{split}
\end{align*}
In the particular case where $\B{u}=\hat{\B{w}} - \B{w}^*$, we obtain with probability at least $1 - \delta /2$:
\begin{equation}\label{lemma2-3}
\frac{1}{n} \sum_{i=1}^n \left(\B{x}_i^T (\hat{\B{w}} - \B{w}^*) \right)^2 \ge 
\| \hat{\B{w}} - \B{w}^* \|_{\infty}^2  e^{- 2\lambda} -  17 \lambda^2 \sqrt{\frac{\log(ep /k)}{n} \log(2/ \delta}).
\end{equation}
Pairing Equation \eqref{lemma2-3} with Lemma \ref{first-lemma-logreg}, and assuming $\lambda \le e^{\lambda}$, we conclude that with probability at least $1- \delta$, it holds uniformly for all the $p$ nodes of the Ising graph:
\begin{align*}
\begin{split}
\| \hat{\B{w}} - \B{w}^* \|_{\infty}^2
&\le  17 \lambda^2 e^{2 \lambda} \sqrt{\frac{\log(ep /k)}{n} \log(2/ \delta})
 + 34 \lambda e^{4 \lambda} \sqrt{\frac{\log(ep(p-1) /k) }{n} \log(2 / \delta)}\\
 &\le  51 \lambda e^{4 \lambda} \sqrt{\frac{\log(e p(p-1) /k) }{n} \log(2 / \delta)}.
\end{split}
\end{align*}
We can consequently conclude that the L0-L2 logistic regression estimator satisfies with probability at least $1 - \delta$:
\begin{equation*}
\| \hat{\B{W}}_{\LR} - \B{W}^* \|_{\infty}^2
\le  51 \lambda e^{4 \lambda} \sqrt{\frac{\log(e p(p-1) /k) }{n} \log(2 / \delta)}.
 \end{equation*}

 \subsection{Proof of Lemma \ref{first-lemma-ise}}\label{sec:first-lemma-ise}

We fix $j \in \{1,\ldots,p\}$ and drop the dependency upon $j$ in this section. The L0-L2 empirical interaction screening estimator $\hat{\B w}$ at node $j$ is defined as a minimizer of the empirical interaction screening loss:
\begin{equation}\label{empirical-problem}
\hat{\B w} ~~ \in \argmin \limits_{ \B{w} \in \mathcal{B}_{0, 2}(k, \lambda) } \mathcal{J}_n(\B{w}) ~~ \text{where} ~~  \mathcal{J}_n(\B{w}) = \frac{1}{n} \sum_{i=1}^n \exp (- y_i \B{x}_i^T  \B{w}).
\end{equation}
We introduce the remainder of the first-order Taylor expansion of $\mathcal{J}_n$ around $\B{w}^*$ defined for every $\B{w} \in \mathbb{R}^{p-1}$ as:
\begin{align*}
\begin{split}
\Delta(\B{w}^*, \B{w}) 
&= \mathcal{J}_n(\B{w}) - \mathcal{J}_n(\B{w}^*) - \nabla \mathcal{J}_n(\B{w}^*)^T(\B{w} - \B{w}^*)\\
&= \frac{1}{n} \sum_{i=1}^n \exp (- y_i \B{x}_i^T  \B{w}^*) 
\left( \exp (- y_i \B{x}_i^T  (\B{w} - \B{w}^*)) - 1 + y_i \B{x}_i^T  (\B{w} - \B{w}^*) \right).
\end{split}
\end{align*}
Let us assume that $\B{w} \in \mathcal{B}_1(\lambda)$. The proof of Lemma 5, \cite{vuffray2016interaction} assess that:
$$e^{-z} - 1 + z \ge \frac{z^2}{2 + |z|}, ~ \forall z \in \mathbb{R}.$$
Because $|y_i \B{x}_i^T  (\B{w} - \B{w}^*)| \le \| \B{w} - \B{w}^* \|_1 \le 2 \lambda$, it consequently holds:
\begin{align}\label{ise-part1}
\begin{split}
\Delta(\B{w}^*, \B{w}) 
&\ge \frac{1}{n} \sum_{i=1}^n \exp (- y_i \B{x}_i^T  \B{w}^*) \frac{1}{2 + \| \B{w} - \B{w}^* \|_1} \left(\B{x}_i^T  (\B{w} - \B{w}^*) \right)^2 \\
&\ge \frac{e^{- \lambda}}{n(2 + 2\lambda)} \sum_{i=1}^n \left(\B{x}_i^T (\B{w} - \B{w}^*) \right)^2.
\end{split}
\end{align}
We now upper-bound the quantity $\Delta(\B{w}^*, \B{w}) $ in the particular case where $\B{w} = \hat{\B{w}}$.  Because $\hat{\B{w}}$ is a minimizer of $\mathcal{J}_n$ satisfying $\| \hat{\B{w}} \|_1 \le \lambda$, it holds:
\begin{align*}
\begin{split}
\Delta(\B{w}^*, \hat{\B{w}} )
& =\mathcal{J}_n(\hat{\B{w}}) - \mathcal{J}_n(\B{w}^*) - \nabla \mathcal{J}_n(\B{w}^*)^T(\hat{\B{w}} - \B{w}^*)\\
&\le - \nabla \mathcal{J}_n(\B{w}^*)^T(\hat{\B{w}} - \B{w}^*)\\
&\le \frac{1}{n} \sum_{i=1}^n y_i \exp (- y_i \B{x}_i^T  \B{w}^*) \B{x}_i^T (\hat{\B{w}} - \B{w}^*)\\
&\le \frac{1}{\sqrt{n}} \sum_{\ell=1}^{p-1} \left\{ \frac{1}{\sqrt{n}} \sum_{i=1}^n y_i \exp (- y_i \B{x}_i^T  \B{w}^*)  x_{i\ell} \right\} h_{\ell}.
\end{split}
\end{align*}
By noting $\tilde{\beta}_{\ell}=\frac{1}{\sqrt{n}} \sum_{i=1}^n y_i \exp (- y_i \B{x}_i^T  \B{w}^*)  x_{i\ell}$, we observe that $\tilde{\beta}_{\ell}=\beta_{j \ell}$, where $\beta_{j \ell}$ has been defined in Lemma \ref{event-C}. Similarly to Equation \eqref{lemma1-second-part}, we assume the event $\mathcal{C}$ defined in Lemma \ref{event-C} is satisfied. We then conclude that with probability at least $1 - \delta/2$:
\begin{align}\label{ise-part2}
\begin{split}
\Delta(\B{w}^*, \hat{\B{w}}) 
&\le 12 \sqrt{\frac{\log(2/ \delta)}{n}} e^{\lambda} \sum_{j=1}^{2k} \lambda_j^{(p(p-1))} |h_j| \\
&\le 34 \sqrt{\frac{\log(2/ \delta)}{n}} \lambda e^{\lambda}  \sqrt{\frac{\log(e p(p-1) / k ) }{n} \log(2 / \delta)}.
\end{split}
\end{align}
Pairing Equations \eqref{ise-part1} and \eqref{ise-part2} we conclude that with  probability at least $1 - \delta/2$:
$$
\frac{1}{n} \sum_{i=1}^n \left(\B{x}_i^T (\hat{\B{w}} - \B{w}^*) \right)^2
\le 68\lambda (1 + \lambda) e^{2\lambda} \sqrt{\frac{\log(e p(p-1) / k ) }{n} \log(2 / \delta)}.
$$

\subsection{Proof of Theorem \ref{upper-bound-ise}}\label{sec:lemma-upper-bound-ise} 
By pairing Lemmas \ref{first-lemma-ise}, \ref{supremum} and \ref{lower-bound}, we derive Theorem \ref{upper-bound-ise} similarly to Section \ref{sec:upper-bound-logreg}.

\bibliography{aaai}